\documentclass[10pt,journal,compsoc]{IEEEtran}
\ifCLASSOPTIONcompsoc
  \usepackage[nocompress]{cite}
\else
  \usepackage{cite}
\fi
\usepackage{cite}
\usepackage{amsmath,amssymb,amsfonts}

\usepackage{graphicx}
\usepackage{textcomp}
\usepackage{xcolor}
\usepackage{makecell}
\usepackage{multirow} 
\usepackage{arydshln} 

\usepackage{algorithm}
\usepackage{algorithmicx}
\usepackage{algpseudocode}  
\usepackage{amsmath}  
\usepackage{amssymb}
\usepackage{ulem}
\usepackage{booktabs} 
\ifCLASSINFOpdf
\else
\fi
\begin{document}
\title{Improving Generalizability of Graph Anomaly Detection Models via Data Augmentation}

% \author{Michael~Shell,~\IEEEmembership{Member,~IEEE,}
%         John~Doe,~\IEEEmembership{Fellow,~OSA,}
%         and~Jane~Doe,~\IEEEmembership{Life~Fellow,~IEEE}% <-this % stops a space
\author{Shuang~Zhou,~\IEEEmembership{Student Member,~IEEE,}
        Xiao~Huang,
        Ninghao~Liu,
        Huachi~Zhou,
        Fu-Lai~Chung,
        and~Long-Kai~Huang% <-this % stops a space
% \IEEEcompsocitemizethanks{\IEEEcompsocthanksitem M. Shell was with the Department
% of Electrical and Computer Engineering, Georgia Institute of Technology, Atlanta,
% GA, 30332.\protect\\
% % note need leading \protect in front of \\ to get a newline within \thanks as
% % \\ is fragile and will error, could use \hfil\break instead.
% E-mail: see http://www.michaelshell.org/contact.html
% \IEEEcompsocthanksitem J. Doe and J. Doe are with Anonymous University.}% <-this % stops an unwanted space
\IEEEcompsocitemizethanks{
\IEEEcompsocthanksitem Shuang Zhou, Xiao Huang, Huachi Zhou, and Fu-Lai Chung are with the Department of Computing, The Hong Kong Polytechnic University, Hong Kong. E-mail: \{csszhou,xiaohuang,cskchung\}@comp.polyu.edu.hk, huachi.zhou@connect.polyu.hk.
\IEEEcompsocthanksitem Ninghao Liu is with the University of Georgia, Georgia, USA. E-mail: ninghao.liu@uga.edu.
\IEEEcompsocthanksitem Long-Kai Huang is with Tencent AI Lab, Shenzhen, China. E-mail: hlongkai@gmail.com.
}

% \thanks{Manuscript received April 19, 2005; revised August 26, 2015.}
}


\markboth{Journal of \LaTeX\ Class Files,~Vol.~14, No.~8, August~2015}%
{Shell \MakeLowercase{\textit{et al.}}: Bare Demo of IEEEtran.cls for Computer Society Journals}



\IEEEtitleabstractindextext{%
\begin{abstract}

% TKDE requires the abstract within 200 words. 
Graph anomaly detection (GAD) has wide applications in real-world networked systems. In many scenarios, people need to identify anomalies on new (sub)graphs, but they may lack labels to train an effective detection model. Since recent semi-supervised GAD methods, which can leverage the available labels as prior knowledge, have achieved superior performance than unsupervised methods, one natural idea is to directly adopt a trained semi-supervised GAD model to the new (sub)graphs for testing. However, we find that existing semi-supervised GAD methods suffer from poor generalization issues, i.e., well-trained models could not perform well on an unseen area (i.e., not accessible in training) of the graph.
Motivated by this, we formally define the problem of generalized graph anomaly detection that aims to effectively identify anomalies on both the training-domain graph(s) and the unseen test graph(s).
Nevertheless, it is a challenging task since only limited labels are available, and the normal data distribution may differ between training and testing data.
Accordingly, we propose a data augmentation method named \textit{AugAN} (\uline{Aug}mentation for \uline{A}nomaly and \uline{N}ormal distributions) to enrich training data and adopt a customized episodic training strategy for learning with the augmented data. Extensive experiments verify the effectiveness of \textit{AugAN} in improving model generalizability.
\end{abstract}

}

\maketitle

% %%%%%%%%%%%%%%%%%%%%%%%%%%%%%%% Introduction %%%%%%%%%%%%%%%%%%%%%%%%%%%%%%%%%%%%%%%

\IEEEraisesectionheading{\section{Introduction}\label{sec:introduction}}
\IEEEPARstart{G}{raph} anomaly detection (GAD) is a crucial task that aims to identify nodes that present strange behaviors and deviate from corresponding normal background significantly~\cite{akoglu2015graph, ma2021comprehensive}. In real-world networked systems, such as social networks~\cite{tan2019deep} and payment transaction graphs~\cite{wang2019semi, chen2022gccad}, even a few anomalies, e.g., financial frauds, may cause tremendous loss.  
Based on feature compactness, anomalies roughly fall into two types~\cite{liu2010detecting, pang2021deep}: scattered anomalies (a.k.a., point anomalies) and rare categories (a.k.a., group anomalies). 
Scattered anomalies are individual instances that randomly appear in feature space and deviate from the majority of other individual samples \cite{xu2019mix}. 
Whereas, rare categories denote some minority groups of data objects that exhibit compact properties in feature space (i.e., share similar behaviors or patterns) and deviate significantly from the vast majority as a whole \cite{han2011data, zhou2018sparc}.
Detecting these anomalies from graphs can eliminate potential loss in real-world scenarios and is of great significance.

\begin{figure}[t]
\begin{center}
\begin{tabular}{c}
\includegraphics[width = 0.95\linewidth]{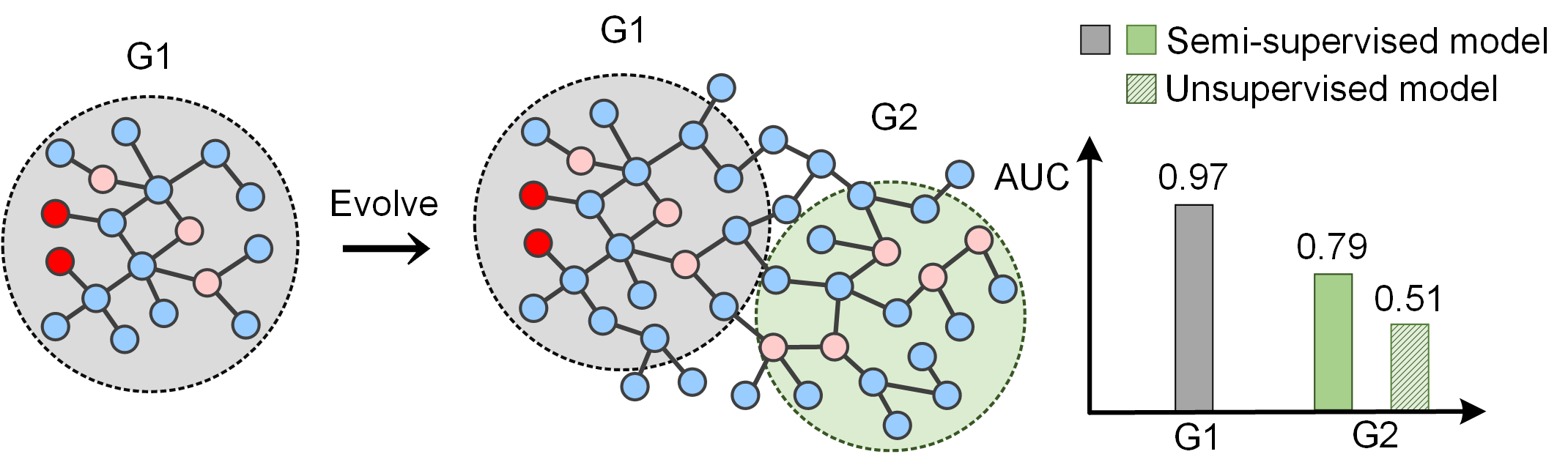}
\end{tabular}
\end{center}
\caption{An empirical study of the generalization issue of GAD models. Although the semi-supervised GAD model trained on graph G1 can effectively detect anomalies on the training-domain graph, its performance severely drops when being tested on an unlabeled graph G2. Note that its performance is still better than that of unsupervised GAD models. (Red color and blue color denote anomaly and normal nodes, respectively. Deep red nodes denote labeled anomalies.)}
\label{Fig1_overfitting_issue}
\vspace{-0.3cm}
\end{figure}

Previous works for anomaly detection on attributed graph are mainly unsupervised methods, including community analysis methods~\cite{gao2010community, gutierrez2020multi}, subspace-selection methods~\cite{sanchez2014local, perozzi2014focused}, residual-based methods~\cite{li2017radar, peng2018anomalous}, and deep learning methods~\cite{peng2020deep, huang2021hybrid, bandyopadhyay2020outlier, jin2021anemone}. 
Because unsupervised methods usually suffer from unsatisfied detection accuracy~\cite{liu2022benchmarking, han2022adbench}, many semi-supervised GAD methods, which can effectively leverage the valuable labeled information as prior knowledge, have been proposed and achieved great performance on the training-domain graph with the condition that sufficient labels are available~\cite{ding2021few, guo2022learning, tang2022rethinking}. 
In practice, graph-based systems would usually evolve and expand to new regions (e.g., states or countries), and people need to identify anomalies on the new (sub)graphs, such as newly-joined communities in social media, traffic networks of new cities, and extended areas in sensor networks. For instance, after PayPal expands its market from the U.S. to Europe, it is necessary to detect frauds on the new (sub)graph formed by European users to secure its business.
But they may lack labels to train an effective detection model, since existing labels are rare and unevenly distributed in large graph and label annotation is costly and time-consuming. One natural idea is to directly adopt a well-trained semi-supervised GAD model to the new (sub)graphs for testing.

But in our pilot study, we find that existing semi-supervised GAD methods suffer from poor generalization issues, i.e., well-trained GAD models (e.g., GDN~\cite{ding2021few}) do not perform well on an unseen area (not accessible in training) of the graph (see Fig.~\ref{Fig1_overfitting_issue}). 
The poor generalization issue may raise great troubles and cannot be ignored because (1) the trained GAD models would easily omit anomalies on the interested yet unlabeled areas, e.g., a newly extended graph; (2) more importantly, the GAD models are not truly ``well-trained" and may lose their effectiveness if the training-domain graph changes (attributes or structure) as time goes on. 
Therefore, we propose to tackle a general and important research problem: \textit{How to learn a generalized GAD model that can effectively identify anomalies on both the training-domain graph(s) and unseen test graph(s)?}

Generalized graph anomaly detection is a non-trivial task due to two reasons. 
First, training-domain graph is usually weakly-annotated with limited labels, which could not provide sufficient supervised information to learn a generalized model.
Second, the normal background (i.e., normal data distribution or normal patterns) usually presents distribution shifts between the training and test graph, and trained GAD models may overfit the normal background in training and could not generalize well to test data.
Inspired by that data augmentation can enrich training data and has been proven effective in improving generalization on related tasks~\cite{wang2022generalizing},
we think that enlarging the quantity of anomalies and enriching the diversity of normal data distributions, by fully leveraging the available training-domain graph(s) with limited labels, is promising to boost the generalizability of semi-supervised GAD models.

However, developing such an intuitive solution based on data augmentation has to tackle two challenges.
The first one is how to augment anomalies on attributed graph. 
Although there exist many node-level data augmentation methods for graphs~\cite{ding2022data, yu2022graph}, including node/edge dropping~\cite{wang2020nodeaug}, graph component synthesis~\cite{park2021graphens}, and feature interpolation~\cite{zhao2021graphsmote, wu2021graphmixup}, only a few of them can handle the scarcity of labeled anomalies in GAD.
Besides, most of the methods are for node classification tasks and
are unsuitable for GAD. Because they generally rely on the ``cluster” assumption~\cite{ruff2019deep} which does not hold on anomaly detection (i.e., anomalies may present diverse characteristics but share the same label).
Second, how to enrich the normal data distributions such that GAD models can effectively identify anomalies under different normal backgrounds (with distribution shifts) is another challenge.
Although a widely-used method called \textit{DeepAll}~\cite{liu2020shape}, which directly combines all the training data for model learning to boost generalization, helps to enrich normal background, it may lead to sub-optimal performance. Because GAD models tend to overfit on the training data and still cannot handle the new background on unseen test data.

To address the aforementioned challenges, we propose a customized data augmentation method named \textit{AugAN}, which consists of anomaly augmentation and normal distribution augmentation, to improve the generalizability of semi-supervised GAD models.
Specifically, we first leverage a shared graph encoder to map all the training graphs into the same latent space to obtain node representations. 
Then, we conduct anomaly augmentation to generate pseudo-labels (i.e., enlarge the anomaly quantity), by inferring those unlabeled nodes as anomalies, if they share similar characteristics to labeled anomalies in other training graphs.
Furthermore, we perform normal distribution augmentation to enable GAD models resistant to the distribution shifts between training and test data. It firstly iteratively masks a random proportion of node representations from the training-domain graphs to enrich the diversity of normal backgrounds,
and then adopts a tailored episodic training strategy, which simulates anomaly detection under different normal backgrounds in training,
to boost GAD model generalizability.
We summarize our contributions as follows.
\begin{itemize}
\item \textbf{Problem:}
We are the first to investigate and formally define the problem of generalized graph anomaly detection. In particular, we simulate real-world scenarios and emphasize boosting detection accuracy on both the training-domain graph(s) and the unseen test graph(s).

% \footnote{{\color{blue}The code and datasets will be released after acceptance.}}
\item \textbf{Algorithm:} We propose a principled data augmentation method \textit{AugAN} to boost GAD model generalizability. 
It enlarges anomaly quantity and enriches the diversity of normal backgrounds, and adopts a tailored episodic training strategy for learning a generalized model with the augmented data. 

\item \textbf{Dataset:}
We release customized datasets for generalization evaluation which also helps to facilitate the development of the generalized GAD.

\item \textbf{Experimental Findings:} We perform extensive experiments to verify the effectiveness of our method for generalized GAD.
Besides, we also discuss many other aspects, including how the number of labels affects generalization, how each component of \textit{AugAN} contributes to its overall performance, and the working mechanism of \textit{AugAN}.

\end{itemize}
% %%%%%%%%%%%%%%%%%%%%%%%%%%%%%%%%%%%%%%%%%%%%%%%%%%%%%%%%%%%%%%%%%%%%%%%%%%%%%

% %%%%%%%%%%%%%%%%%%%%%%%%%%%%%% Problem Statement %%%%%%%%%%%%%%%%%%%%%%%%%%

\section{Problem Statement}
\begin{figure*}
\begin{center}
\includegraphics[width = 1.0\linewidth]{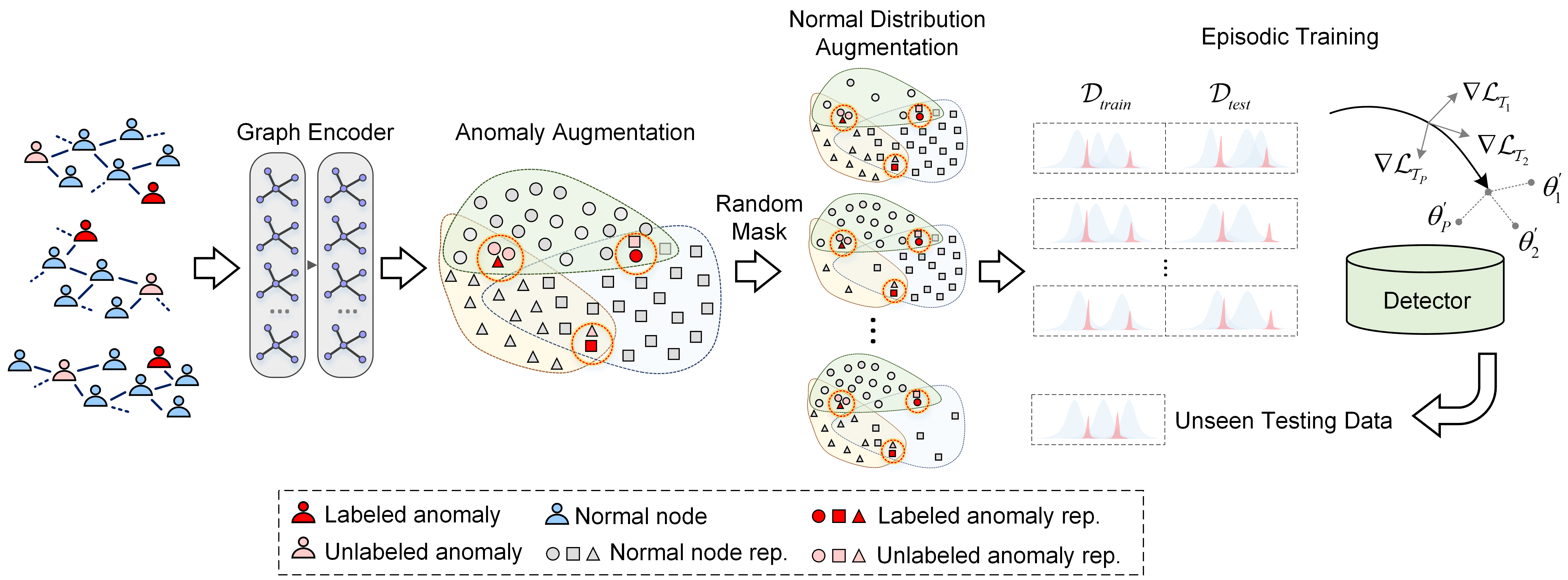}
\end{center}
\caption{Overview of the proposed method \textit{AugAN}. It first leverages a shared graph encoder to obtain node representations. Then, it performs anomaly augmentation to enlarge the quantity of labeled anomalies. To enable identifying anomalies under various normal backgrounds, it conducts normal distribution augmentation, and employs a tailored episodic training for model learning.
}
\label{Fig2_AID_framework}
\vspace{-0.2cm}
\end{figure*}

In this paper, we use bold uppercase letters (e.g., \textbf{X}), bold lowercase letters (e.g., \textbf{h}), lowercase letters (e.g., $s$), and calligraphic fonts (e.g., $\mathcal{V}$) to denote matrices, vectors, scalars, and sets, respectively.
We represent an attributed graph as $\mathcal{G} = (\mathcal{V}, \mathcal{E}, \textbf{X})$,
where $\mathcal{V}=\{v_1$, $v_2$, ..., $v_n\}$ is the set of nodes and $\mathcal{E}$ is the set of edges. 
The node attributes are represented by \textbf{X} = [$\textbf{x}_1$, $\textbf{x}_2$, ..., $\textbf{x}_n$] $\in \mathbb{R}^{n \times d}$, $\textbf{x}_i \in \mathbb{R}^{d}$ is the attribute vector associated with node $v_i$ and $d$ is the attribute dimension. 

\noindent\textbf{Definition: Graph Anomaly Detection.}\footnote{In this paper, we mainly focus on detecting abnormal nodes, especially rare categories.} Given an attributed graph, finding the nodes that present different patterns and deviate from the vast majority (i.e., normal background) significantly. The output is usually a ranking list ordered by anomaly score where anomalies own higher anomaly scores than normal data and are ranked top.

Generally speaking, generalized graph anomaly detection aims to maximally improve the detection performance on unlabeled graph(s) through employing very limited ground-truth anomalies from training graph(s). 
In this work we assume there exist \textit{m} (\textit{m}$\geq$2) training-domain graphs for two reasons: (1) In real-world scenarios, the training graphs may refer to different but related networked systems, e.g., LinkedIn and Indeed~\cite{ding2021few}; (2) In practice, a large graph is usually divided into different sub-graphs for various reasons, e.g., region. For example, the transaction graph in PayPal can be naturally partitioned into different sub-graphs formed by users in Europe, the U.S., etc. 
Formally, the \textit{m} training-domain graphs $\mathcal{G}^L = \{\mathcal{G}^L_1, \mathcal{G}^L_2, ..., \mathcal{G}^L_m\}$ (\textit{m}$\geq$2) are related with unlabeled testing graphs $\mathcal{G}^U = \{\mathcal{G}^U_1, \mathcal{G}^U_2, ..., \mathcal{G}^U_e\}$  (\textit{e}$\geq$1), but these graphs may exist distribution shifts.
For each graph in $\mathcal{G}^L$, the set of labeled anomalies is denoted as $\mathcal{S}^L$ and the set of unlabeled nodes is represented as $\mathcal{V}^U$, i.e., $\mathcal{V}$ = $\left\{\mathcal{S}^L, \mathcal{V}^U\right\}$. Note that, $\left| \mathcal{S}^L \right| \ll \left| \mathcal{V}^U \right| $, since only a few labeled anomalies are available in practice.
Further, some nodes are randomly sampled from $\mathcal{V}^U$ and regarded as ``normal" data for training (i.e., $\mathcal{V}_{norm}$), while the rest of $\mathcal{V}^U$ is used for testing (i.e., $\mathcal{V}_{test}$), thus 
$\mathcal{V}$ = $\left\{\mathcal{S}^L, \mathcal{V}_{norm},  \mathcal{V}_{test}\right\}$.
For each graph in $\mathcal{G}^U$, the whole graph is used for testing.

\noindent\textbf{Problem: Generalized Graph Anomaly Detection.}
Given some graphs $\mathcal{G}^L = \{\mathcal{G}^L_1, \mathcal{G}^L_2, ..., \mathcal{G}^L_m\}$, each of which contains a few labeled anomalies (i.e.,$\mathcal{S}^L_1, \mathcal{S}^L_2, ..., \mathcal{S}^L_m$),
our goal is to train a generalized detection model that can effectively detect anomalies on both the $\mathcal{V}_{test}$ of the training-domain graphs and the unlabeled (unseen) graphs $\mathcal{G}^U$. 

% %%%%%%%%%%%%%%%%%%%%%%%%%%%%%%%%%%%%%%%%%%%%%%%%%%%%%%%%%%%%%%%%%%%%%%%%%%%%%

% %%%%%%%%%%%%%%%%%%%%%%%%%%%%%% Method %%%%%%%%%%%%%%%%%%%%%%%%%%%%%%%%%%
\section{Methodology}
In this paper, we propose a customized data augmentation method called \textit{AugAN}, which consists of anomaly augmentation and normal distribution augmentation. 
Specifically, we first leverage a shared graph encoder to map all the training graphs into a latent space.
To enlarge the quantity of labeled anomalies for learning generalized models, we conduct \textbf{anomaly augmentation} to generate pseudo-labels.
To enable GAD models effectively identify anomalies under various normal backgrounds with distribution shifts, we perform \textbf{normal distribution augmentation} and further propose a tailored episodic training strategy to handle unseen backgrounds in test graphs. An overview of the method is depicted in Fig.~\ref{Fig2_AID_framework}.

\subsection{Graph Encoder}
To effectively leverage the graph structure and node attributes in training data for learning generalized models, we build a shared graph encoder to map all the training graphs into the same latent space.
Specifically, the graph encoder consists of multiple graph neural network (GNN) layers that encode nodes to low-dimensional representations. Based on the message-passing mechanism, each GNN layer updates node representations by aggregating information from neighborhoods.
Thus, the representations after $k$ layers' aggregation would capture the information of the $k$-hop neighborhoods on graph.
Formally, the updating process of the $k$-th layer in GNN is formally represented as:
\begin{equation}
\setlength{\abovedisplayskip}{3pt}
\setlength{\belowdisplayskip}{3pt}
\begin{array}{lll}
\textbf{a}_{i}^{(k)}=\textsc{Aggregate}^{(k-1)}\left(\textbf{h}_{j}^{(k-1)}, {\forall} v_j \in \mathcal{N}_i\right), \\
\textbf{h}_{i}^{(k)}=\textsc{Combine}^{(k)}\left(\textbf{h}_{i}^{(k-1)}, \textbf{a}^{(k)}\right),
\end{array}
\label{Eq_GNN_genral}
\end{equation}
where $\mathbf{h}_{i}^{(k)}$ is the representation of $v_i \in \mathcal{V}$ at $k$-th layer, and $\mathcal{N}_i$ is a set of neighborhoods of $v_i$. 
Note that the same graph encoder is employed on all the training graphs to encode them into a shared latent space.

\subsection{Anomaly Augmentation}
The goal of anomaly augmentation is to enlarge anomaly quantity in training data for learning a generalized detection model.
Although some recent graph-based data augmentation methods~\cite{zhao2021graphsmote, wu2021graphmixup} can alleviate the scarcity of labels by randomly interpolating labeled anomalies' representations to generate pseudo-labels, directly applying these methods may lead to the generated features not denoting abnormal patterns and is detrimental to model learning.
Accordingly, we propose to fully leverage the training graphs and focus on discovering nodes with similar characteristics to the known anomalies to generate pseudo-labels for anomaly augmentation.
Specifically, we assume that 
some anomalies shared with similar characteristics are scattered among the training graphs, but only a few of them are labeled. Hence, we may infer those unlabeled nodes with similar characteristics to any labeled anomalies from other graphs as anomalies, and then leverage these high-confidence nodes to generate synthetic representations as pseudo-labels, thus enlarging anomaly quantity in training.

Formally, for an anomaly $v_a^{(i)}$ from $\mathcal{G}^L_i$, we leverage it to augment anomalies via two steps. First, we build the high-confidence node set $S^{h}$ from all the external graphs via 
\begin{equation}
\begin{aligned}
\mathcal{S}^{h}=\left\{ v_b^{(j)} \in \mathcal{G}_{j}^{L} \vert  \mathcal{R}\left(v_a^{(i)}, v_b^{(j)}\right)<\eta\right\},
% \text {and } \mathcal{D}\left(v_{p}^{m}, v_{i}^{n}\right)<\eta\right\},
\end{aligned}
\label{Eq_1_external}
\end{equation}
where node $v_b^{(j)}$ is from an external graph $\mathcal{G}_{j}^{L}$ (i.e., $i\neq j$), $\mathcal{R}\left(\cdot,\cdot\right)$ denotes a distance function to measure nodes similarity in latent space, and we use Euclidean distance, i.e., 
\begin{equation}
\mathcal{R}\left(v_a^{(i)}, v_b^{(j)}\right) = \Vert \textbf{h}_a^{(i)} - \textbf{h}_b^{(j)}\Vert^{2}_{2},
\label{Eq_2_euclidean}
\end{equation}
where $\textbf{h}_a^{(i)}$ and $\textbf{h}_b^{(j)}$ denote the representation of node $v_a^{(i)}$ and $v_b^{(j)}$.
The $\eta$ is a threshold which requires that the selected nodes from $\mathcal{G}_{j}^{L}$ should be closer to the target node $v_a^{(i)}$ than any other nodes in $\mathcal{G}^L_i$, i.e.,
\begin{equation}
\eta = \sigma \cdot \operatorname{min} \mathcal{R}\left(v_{a}^{(i)}, v_{c}^{(i)}\right), \forall v_{c}^{(i)} \in \mathcal{G}^L_i,
\label{Eq_3_threshold}
\end{equation}
where $\sigma \in (0, 1) $ is the weight to control $\eta$. 
Second, we base on the high-confidence anomalies to generate synthetic representations as pseudo-labels via
\begin{equation}
\textbf{h}_{new}=(1-\lambda) \cdot \textbf{h}_{a}^{(i)}+\lambda \cdot \textbf{h}_{b}^{(j)},
\label{Eq_data_aug_SMOTE}
\end{equation}
where $\lambda$ is a random variable in the range of (0, 1), and we can base on a predefined augmentation factor $\alpha$ to generate multiple pseudo-labels by varying the value of $\lambda$. In short, for each labeled anomaly, we try to find its corresponding $\mathcal{S}^h$ and generate $\textbf{h}_{new}$ as pseudo-labels, and we use $\mathcal{S}^{r}$ to denote all the generated pseudo-labels.

\subsection{Normal Distribution Augmentation}
After anomaly augmentation, we conduct normal distribution augmentation to create diversified backgrounds with different normal distributions. 
The main idea is that we iteratively mask a random proportion of nodes from all the training data to boost the diversity of normal backgrounds, and further perform a customized episodic training to help GAD models alleviate overfitting to training data and handle unseen background in the unseen test graphs.

Specifically, we implement the normal distribution augmentation by two steps. First, we merge all the anomalies and normal data respectively from the available training graphs via 
\begin{equation}
\begin{aligned}
& \mathcal{S}_{merge}=\left\{v_{p}^{(i)} \in \mathcal{S}_{i}^{L} \cup \mathcal{S}^{r}, i \in 1,2,..., m\right\}, \\
&\mathcal{N}_{merge}=\left\{v_{q}^{(i)} \in \mathcal{V}_{norm}^{(i)}, i \in 1,2,..., m\right\},
\end{aligned}
\label{Eq_7_merge_data}
\end{equation}
where $\mathcal{N}_{merge}$ and $\mathcal{S}_{merge}$ denote the merged node set for normal data and labeled anomalies, respectively. 
Second, we iteratively and randomly mask a proportion of nodes from $\mathcal{N}_{merge}$ in the latent space to obtain $ \mathcal{N}_{mask}$, thus creating numerous backgrounds with diverse normal distributions. Here, we call a newly generated background together with the anomalies as a \textit{scene}. Formally, each \textit{scene} $\mathcal{C}_{i}$ can be represented as $\mathcal{C}_{i}=\left\{\mathcal{S}_{merge}, \mathcal{N}_{mask}\right\}$.
% \begin{equation}
% \begin{aligned}
% &\mathcal{C}_{i}=\left\{\mathcal{S}_{merge}, \mathcal{N}_{mask}\right\}.
% \end{aligned}
% \label{Eq_8_scene_Ci}
% \end{equation}
The generated \textit{scenes} in training data greatly enhance the diversity of normal distribution and support the subsequent episodic training.

\begin{algorithm}[htb]  
\small
\caption{\textit{AugAN} Equipped GAD Model Learning.} 
\label{alg_framwork_data_diversify}  
\begin{algorithmic}[1]  
\Require  
(1) training graphs, $\mathcal{G}^L = \{\mathcal{G}^L_1, \mathcal{G}^L_2, ..., \mathcal{G}^L_m\}$ (\textit{m}$\geq$2);
(2) a few labeled anomalies for $\mathcal{G}^L$ (i.e.,$\mathcal{S}^L_1, \mathcal{S}^L_2, ..., \mathcal{S}^L_m$);
(3) training epochs \textit{E}, task number \textit{P}, batch size $t$, augmentation factor $\alpha$, weight value $\sigma$, mask ratio $\rho$, and learning rate $r_{1}, r_{2}$;
(4) a base detection model;
(5) testing graph $\mathcal{G}^{U}$.
\Ensure  
Anomaly scores of nodes for $\mathcal{G}^{L}$ and $\mathcal{G}^{U}$.

\State Randomly initialize graph encoder;
\label{fram:1}  
 
\State For each anomaly $v_a^{(i)}$ from $\mathcal{G}_{i}^{L}$, select high-confidence node set $\mathcal{S}^{h}$ from all the external graphs based on Eqs. \eqref{Eq_1_external} - \eqref{Eq_3_threshold};  
\label{fram:2}  

\State Generate virtual representation $\textbf{h}_{new}$ based on Eq. \eqref{Eq_data_aug_SMOTE} and $\alpha$, and get all the pseduo-labels $\mathcal{S}^{r}$;
\label{fram:3}

\While {epoch $< \textit{E}$}

\State   Merge all the training data via Eq. \eqref{Eq_7_merge_data};

\State   Base on $\mathcal{C}_{i}=\left\{\mathcal{S}_{merge}, \mathcal{N}_{mask}\right\}$ to randomly mask data 
\Statex  \quad\;  with ratio $\rho$ to create numerous \textit{scenes};
% \State   Draw \textit{P} tasks from the \textit{scenes};

\For {each task $\mathcal{T}_{i}$}

\State   Randomly sample $\frac{t}{2}$ nodes from $\mathcal{S}_{merge}$ and $\mathcal{N}_{mask}$
\Statex  \qquad\quad of \textit{scene} $\mathcal{C}_i$ to form a balanced set $\mathcal{D}_{train}$;

\State  Evaluate $\nabla_{\theta} \mathcal{L}_{\mathcal{D}_{train}}\left(f_{\theta}\right)$;
\State  Compute parameters $\theta^{\prime}$ based on Eq. \eqref{Eq_9_theta_prime};
\State  Sample a balanced set $\mathcal{D}_{test}$ from a different \textit{scene} $\mathcal{C}_j$
\Statex  \qquad\quad  for the meta-update;

\EndFor 
\State  Update Eq. \eqref{Eq_11_meta_optimization_theta} using $\mathcal{D}_{test}$;
\EndWhile

\State Compute anomaly scores for $\mathcal{G}^{U}$ and testing set of $\mathcal{G}^{L}$. 
\end{algorithmic}  
\end{algorithm}

\subsection{Episodic Training}
We further propose a tailored training strategy to enable GAD models effectively identify anomalies under various normal backgrounds.
To achieve this, during the training process, we iteratively simulate the scenarios of anomaly detection on unseen data, where the model is encouraged to find suitable parameters that generalize well on new testing data with different normal distributions.

Inspired by the meta-learning algorithm~\cite{finn2017model}, which 
aims to achieve fast adaptation with limited labeled data, we propose a \textbf{customized episodic training strategy} for improving GAD model generalizability.
Specifically, we denote a base detection model as $f_{\theta}$ with parameters $\theta$. We use $p(\mathcal{T})$ to denote the distribution over the generated \textit{scenes}, and draw \textit{P} tasks from $p(\mathcal{T})$. Each task $\mathcal{T}_{i}$ contains a task-specific dataset $\mathcal{D}_{T}$ that consists of a support set $\mathcal{D}_{train}$ and a query set $\mathcal{D}_{test}$, i.e.,
$\mathcal{D}_{T}=\left\{\mathcal{D}_{ train}, \mathcal{D}_{test}\right\}$.
% \begin{equation}
% \begin{aligned}
% &\mathcal{D}_{T}=\left\{\mathcal{D}_{ train}, \mathcal{D}_{test}\right\}.
% \end{aligned}
% \label{Eq_9_task_Ti}
% \end{equation}
Note that, $\mathcal{D}_{train}$ and $\mathcal{D}_{test}$ are sampled from two different \textit{scenes}, thus owning different normal distributions. Besides, each support set or query set is comprised of a batch of $t$ data instances with half of the instances from $\mathcal{S}_{merge}$ and another half from $\mathcal{N}_{mask}$.

Briefly, our episodic training (i.e., meta-learning) is conducted via two optimization loops over \textit{P} tasks: (1) outer loop, which updates the neural network parameters in the base detection model on $\mathcal{D}_{test}$ for finding suitable parameters; (2) inner loop, which takes the initialized parameters from outer loop and performs a few gradient updates over $\mathcal{D}_{train}$ to examine model generalizability.
The optimization algorithm adapts for each task $\mathcal{T}_{i}$ independently. 
When adapting to the task $\mathcal{T}_{i}$, the detection model’s parameters $\theta$ are updated to $\theta_{i}^{\prime}$. 
Formally, the parameters update with one gradient step can be described as:
\begin{equation}
\theta_{i}^{\prime}=\theta-r_1 \nabla_{\theta} \mathcal{L}_{\mathcal{D}_{train}}\left(f_{\theta}\right),
\label{Eq_9_theta_prime}
\end{equation}
where $r_1$ is the task-learning rate, $\mathcal{L}$ denotes the corresponding loss function for $f_{\theta}$. 
For simplicity, we only describe one gradient update, and extending it to multiple gradient updates is straightforward.

The base detection model is trained by optimizing for the best performance of $f_{\theta}$ across all the tasks. 
In general, the meta-objective function is as follows:
\begin{equation}
\begin{gathered}
\min _{\theta} \sum_{\mathcal{D}_{test} \sim p(\mathcal{T})} \mathcal{L}_{\mathcal{D}_{test}}\left(f_{\theta_{i}^{\prime}}\right) \\
= \min _{\theta} \sum_{\mathcal{D}_{test} \sim p(\mathcal{T})} \mathcal{L}_{\mathcal{D}_{test}}\left(f_{\theta-r_1 \nabla_{\theta} \mathcal{L}_{\mathcal{D}_{train}}\left(f_{\theta}\right)}\right).
\end{gathered}
\label{Eq_10_meta_overall_goal}
\end{equation}
Note that, the set of instances $\mathcal{D}_{test}$ used for the meta-update is sampled from a different \textit{scene}. 
It is because we aim to simulate the scenarios of anomaly detection on unseen data to find suitable model parameters that can generalize well on new testing data with different normal distribution.

More concretely, the meta-optimization across tasks is performed via gradient descent, such that the model parameters $\theta$ are updated as:
\begin{equation}
\theta \leftarrow \theta- r_2 \nabla_{\theta} \sum_{\mathcal{D}_{test} \sim p(\mathcal{T})} \mathcal{L}_{\mathcal{D}_{test}}\left(f_{\theta_{i}^{\prime}}\right),
\label{Eq_11_meta_optimization_theta}
\end{equation}
where $r_2$ is a learning rate.
The detailed process of the episodic training with the augmented data is presented in Algorithm \ref{alg_framwork_data_diversify}.

% %%%%%%%%%%%%%%%%%%%%%%%%%%%%%%%%%%%%%%%%%%%%%%%%%%%%%%%%%%%%%%%%%%%%%%%%%%%%%

% %%%%%%%%%%%%%%%%%%%%%%%%%%%%%% Exp %%%%%%%%%%%%%%%%%%%%%%%%%%%%%%%%%%
\section{Experiments}

In this section, we perform empirical evaluations on real-world attributed graphs to answer the following research questions:
\textbf{RQ1:} How serious is the poor generalization issue on semi-supervised GAD models and how effective is our method to boost model generalizability? 
\textbf{RQ2:} How effective is our method when only very limited anomaly labels are available?
\textbf{RQ3:} How is the stability of the method?
\textbf{RQ4:} How is the efficiency of the method?
\textbf{RQ5:} What is the contribution of each component in our method?
\textbf{RQ6:} How does our method work in practice?

\subsection{Dataset}
We employ five real-world attributed graphs and further process them to create customized datasets for semi-supervised GAD model generalizability evaluation. Details of the datasets are as follows.

\begin{itemize}

\item \textbf{Clothing} is an attributed graph from Amazon~\cite{mcauley2015inferring}. In the dataset, nodes denote products, and the edge between nodes means the products have ever viewed by the same user. Node attributes are derived from product descriptions. 
The node class is defined by the fine-grained product categories. 

\item \textbf{Photo} is from the Amazon co-purchase graph~\cite{mcauley2015image} in which nodes denote products, and the edges reflect co-purchase relationships. Node attributes are a bag-of-words representation of a product’s reviews.

\item \textbf{ACM} is a citation network from Aminer \cite{tang2008arnetminer} where each node presents a paper, and the edges represent citations. We apply the bag-of-words model to generate node attributes from the abstract and employ the publication venues as node classes.

\item \textbf{DBLP} is a citation network from Aminer \cite{tang2008arnetminer} where nodes represent papers, and edges denote the citations among papers. Node attributes are a bag-of-words representation of the paper abstract. The node class denotes the venues of the paper.

\item \textbf{CS} is a co-author network from the Microsoft Academic Graph~\cite{sinha2015overview}, in which nodes represent authors and the edges indicate co-author relationships. 
Node attributes are a bag-of-words representation of the keywords from an author’s papers.

\end{itemize}

Following the standard anomaly detection settings~\cite{zhou2018sparc, wang2021one, zhou2022unseen}, we convert the existing multi-classification graph datasets (e.g., DBLP) into GAD datasets by treating one or several small classes to be “anomalies” (i.e., rare categories) while the other classes as “normal”.
Further, we are inspired by the related works~\cite{ding2021few, GRDA} to randomly partition a large graph into subgraphs of similar size and create distribution shifts among the subgraphs, which help with model generalizability evaluation~\cite{wang2022generalizing}.
Specifically, we first randomly select a node from the graph and take it as an anchor. Then, we continue to find the second node which is farthest from the first node based on its shortest-path distance. The above process is repeated until we find $m$ nodes that are far away from each other. Next, we take each of the selected nodes as a small subgraph and span them by adding their $k$-hop neighbors into the corresponding subgraphs. Later, we examine the spanned subgraphs and control the overlapped nodes to ensure that there are no overlapped anomalies and the proportion of overlapped normal nodes is less than 10\% in each subgraph. Finally, we save the largest connected component from each of the subgraphs. 
We repeat the above steps multiple times until the generated subgraphs fit all the requirements. In this way, the partitioned subgraphs are assumed to lie on different regions of the large graph and present distribution shifts~\cite{GRDA}.
Table~\ref{Table_1_dataset} summarizes the statistics of the original datasets and processed datasets. 

% \footnote{Refer to the \textit{Experimental Settings} section.}
\begin{table}[]
\caption{Data statistics of the customized datasets for GAD generalizability evaluation. $r$ denotes the anomaly ratio, and $\beta$ denotes the contamination level in training data\protect\footnotemark}.
\resizebox{1.0\linewidth}{!}{
\begin{tabular}{cccccc}
\toprule
\multicolumn{1}{c}{\textbf{Datasets}} & \textbf{Clothing} & \textbf{Photo} & \textbf{ACM} & \textbf{DBLP} & \textbf{CS} \\
\midrule
original nodes          & 24,919              & 7,487            & 22,944        & 40,672          & 18,333        \\
original classes          & 77              & 8            & 9        & 137          & 15        \\
rare categories          & 9              & 2            & 2        & 15          & 3        \\
nodes (avg.)      & 5,568              & 2,736           & 7,564         & 10,014         & 4,713        \\
edges (avg.)      & 17,471             & 37,914          & 31,488        & 25,552         & 18,216       \\
attributes          & 9,034              & 745            & 10,000        & 7,202          & 6,805        \\
anomalies (avg.)  & 170               & 165            & 461          & 534           & 262         \\
$r$ (avg.)          & 3.05\%            & 6.03\%        & 6.09\%       & 5.33\%        & 5.56\%        \\
$\beta$ (avg.)          & 2.16\%            & 4.20\%        & 5.42\%       & 4.82\%        & 4.51\%        \\
\bottomrule
\end{tabular}
}
\label{Table_1_dataset}
\end{table}
\footnotetext{The statistics is based on the setup in \textit{Experimental Settings}.}

\subsection{Experimental Settings}
$\\$
\noindent\textbf{Comparative Methods.}
We compare our proposed method with three categories of methods for GAD generalizability evaluation, including (1) unsupervised GAD methods (Dominant, CoLA, SL-GAD), (2) semi-supervised methods within single graph scenario (OCGNN, GCN, DeepSAD, PCGNN, BWGNN, GDN), and (3) semi-supervised methods for multiple graphs scenario (Commander, GNN-EERM, GDN-MAML). 
Details of these existing baseline methods are as follows:
\begin{itemize}
\item \textbf{Dominant} \cite{ding2019deep} is a GNN-based unsupervised framework that computes anomaly scores by leveraging both network structure and node attributes; it is widely used in related works.

\item \textbf{CoLA} \cite{liu2021anomaly} is an advanced unsupervised GAD method that combines the advantages of graph neural networks (GNNs) and contrastive learning.

\item \textbf{SL-GAD} \cite{zheng2021generative} is a state-of-the-art (SOTA) unsupervised GAD method based on contrastive learning.

\item \textbf{OCGNN} \cite{wang2021one} 
is an advanced GAD method that combines GNNs with a classical one-class objective. 

\item \textbf{GCN} \cite{kipf2017semi} is a popular GNN with cross-entropy loss for node classification, and we adopt it for GAD.

\item \textbf{DeepSAD} \cite{ruff2019deep} 
is a popular feature-based anomaly detection method. Here, we use GNNs as its feature extractor.

\item \textbf{PCGNN} \cite{liu2021pick} 
is a tailored GNN for fraud detection on multi-relation graphs. Here, we adopt it for anomaly detection on attributed graphs.

\item \textbf{BWGNN} \cite{tang2022rethinking} 
is an advanced GNN for semi-supervised GAD.

\item \textbf{GDN}~\cite{ding2021few} 
is a SOTA semi-supervised GAD method that uses established GNNs for feature learning and adopts deviation loss~\cite{pang2019deep} for training.

\item \textbf{Commander}~\cite{ding2021cross} is an advanced method that leverages supervision signals across graphs to boost GAD accuracy on an unlabeled graph. 

\item \textbf{GNN-EERM} \cite{wu2022handling} is a domain-invariant learning method to improve generalization for node classification on graph. Here, we adopt it for GAD.

\item \textbf{GDN-MAML} \cite{ding2021few} is a SOTA GAD method that leverages a few labeled anomalies from multiple graphs; it equips GDN with MAML~\cite{finn2017model} for training.

\end{itemize}
Note that our proposed method can be applied to arbitrary detection models that support episodic training. 
In the experiments, we adopt GCN and GDN as the detection model and further compare our method with (1) a graph-based data augmentation method abbreviated as \textit{SMOTE}~\cite{zhao2021graphsmote}, (2) a strong baseline \textit{DeepAll}~\cite{liu2020shape} (i.e., combining all the training data for model learning), which is a widely-used method on related tasks~\cite{dou2019domain}, (3) a recently proposed data augmentation method called \textit{DAGAD}~\cite{liu2022dagad}, which is applicable to GNN-based GAD methods, for a fair evaluation.

\noindent\textbf{Evaluation Metrics.}
We follow previous works~\cite{liu2021anomaly, ding2021few} and adopt two standard evaluation metrics (AUC and AUPR) to measure the effectiveness of all the methods. 
Specifically, AUC is the area under the ROC curve, which is a plot of true positive rate against false positive rate. AUC score is computed based on the relative ranking of predicted anomaly scores of all instances, which can eliminate the influence of imbalanced classes. 
AUPR is the area under the curve of precision against recall at different thresholds, and it merely evaluates the performance on the positive class (i.e., abnormal samples).

\noindent\textbf{Implementation Details.}
We implement the proposed method in PyTorch. For the employed base detection models (i.e., GCN\footnote{https://github.com/tkipf/pygcn}, PCGNN\footnote{https://github.com/PonderLY/PC-GNN}, GDN\footnote{https://github.com/kaize0409/Meta-GDN\_AnomalyDetection}), we adopt the authors' official codes in our experiments. 
We set the tasks number \textit{P} as 30, set batch size as 128, leverage a 5-step gradient update to compute $\theta_{i}^{\prime}$, and adopt Adam optimizer for GAD model training. We train the GAD models equipped with \textit{AugAN} for 2000 epochs. 
To examine model generalizability, we adopt the widely-used leave-one-out scheme to hold out one graph as unseen graph for testing and adopt the remained graphs for training.
For the training graphs, we further split the nodes on each of them into train, validation, and test set with the proportion of 4:2:4. All the train parts are combined for model learning while all the validation parts are combined for hyper-parameter tuning~\cite{wang2022generalizing}. Notably, the labeled anomalies $\mathcal{S}^L$ and ``normal'' data $\mathcal{V}_{norm}$ are respectively split into train and validation set with the proportion (i.e., 4:2) for model learning and hyper-parameter tuning.
If not further specified, we set the $\left|\mathcal{S}^L\right|$ as 20 for each $\mathcal{G}^l$.

For the other comparative methods, we adopt the authors' official codes for CoLA\footnote{https://github.com/GRAND-Lab/CoLA}, SL-GAD\footnote{https://github.com/yixinliu233/SL-GAD}, OCGNN\footnote{https://github.com/WangXuhongCN/OCGNN}, BWGNN\footnote{https://github.com/squareRoot3/Rethinking-Anomaly-Detection}, and GNN-EERM\footnote{https://github.com/qitianwu/GraphOOD-EERM}. As for Dominant, we leverage the implementation from PyGOD\footnote{https://github.com/pygod-team/pygod}~\cite{liu2022benchmarking}, a comprehensive PyTorch-based library for unsupervised GAD. As for DeepSAD\footnote{https://github.com/lukasruff/Deep-SAD-PyTorch}, we base on its official code and further employ GNNs (e.g., GCN) as its feature extractor to customize it for GAD. Besides, we implement Commander in PyTorch based on the paper description. 
For fair comparison, we select the hyper-parameters with the best performance on the validation set for all the semi-supervised methods. For the unsupervised GAD methods, we follow a previous work~\cite{liu2022benchmarking} and adopt the hyper-parameters that lead to the best performance on the target dataset. We run experiments 10 times and report the averaged results.

\begin{table*}[t]
\centering
\caption{Model performance w.r.t AUC and AUPR on the testing set of training-domain data. \textbf{Bold}: best; \uline{Underline}:runner-up.}
\resizebox{0.75\linewidth}{!}{
\begin{tabular}{ccccccccccc}
\toprule
\multirow{2}{*}{\textbf{Method}} & \multicolumn{2}{c}{\textbf{Clothing}} & \multicolumn{2}{c}{\textbf{Photo}} & \multicolumn{2}{c}{\textbf{ACM}}            & \multicolumn{2}{c}{\textbf{DBLP}}           & \multicolumn{2}{c}{\textbf{CS}}             \\    \cmidrule(r){2-3} \cmidrule(r){4-5} \cmidrule(r){6-7}  \cmidrule(r){8-9}  \cmidrule(r){10-11} 
                        & \textbf{AUC}      & \textbf{AUPR}     & \textbf{AUC}    & \textbf{AUPR}    & \textbf{AUC}         & \textbf{AUPR}        & \textbf{AUC}         & \textbf{AUPR}        & \textbf{AUC}         & \textbf{AUPR}        \\
\midrule
\textbf{Dominant}                & 0.5281            & 0.0379            & 0.4984           & 0.0733          & 0.4810          & 0.0613          & 0.5243          & 0.0717          & 0.4061          & 0.0357          \\
\textbf{CoLA}                    & 0.4635            & 0.0298            & 0.5297           & 0.0741          & 0.4723          & 0.0621          & 0.4645          & 0.0533          & 0.4900          & 0.0634          \\
\textbf{SL-GAD}                  & 0.5720            & 0.0386            & 0.5328           & 0.0850          & 0.5171          & 0.0660          & 0.5336          & 0.0598          & 0.5598          & 0.0731          \\
\midrule
\textbf{OCGNN}                   & 0.6416            & 0.0502            & 0.6077           & 0.0855          & 0.6179          & 0.0751          & 0.5505          & 0.0732          & 0.7607          & 0.4994          \\
\textbf{GCN}                     & 0.7254            & 0.1708            & 0.9120           & 0.6623          & 0.7939          & 0.2895          & 0.6210          & 0.0941          & 0.9127          & 0.5933          \\
\textbf{DeepSAD}                 & 0.9074            & 0.5320            & 0.9308           & 0.5911          & 0.6832          & 0.2201          & 0.7239          & 0.2158          & 0.8628          & 0.4814          \\
\textbf{PCGNN}                   & 0.7198            & 0.1756            & 0.9169           & 0.6716          & 0.7556          & 0.2123          & 0.6019          & 0.0927          & 0.9156          & 0.6008          \\
\textbf{BWGNN}                   & 0.8172            & 0.4021            & 0.9027           & 0.6626          & 0.7945          & 0.2708          & 0.6629          & 0.1113          & 0.9171          & 0.7181          \\
\textbf{GDN}                     & 0.8901            & 0.4981            & 0.9409           & 0.7705          & 0.8631          & 0.3958          & 0.7519          & 0.2947          & 0.9550          & 0.8518          \\
\midrule
\textbf{Commander}               & 0.6471            & 0.0491            & 0.6134           & 0.0830          & 0.5983          & 0.0856          & 0.5774          & 0.0752          & 0.7608          & 0.3034          \\
\textbf{GNN-EERM}                & 0.6911            & 0.2596            & 0.7630           & 0.1958          & 0.7610          & 0.2035          & 0.8222          & 0.2694          & 0.9211          & 0.6231          \\
\textbf{GDN-MAML}                & \textbf{0.9128}   & 0.4898            & \uline{0.9590}           & 0.7807          & \uline{0.8885}          & \textbf{0.4382} & 0.8273          & 0.2602          & \textbf{0.9913} & 0.8978          \\ 
\midrule
\textbf{GCN-DeepAll}             & 0.8760            & 0.4851            & 0.9241           & 0.7057          & 0.7957          & 0.2871          & 0.7171          & 0.2023          & 0.9255          & 0.6464          \\
\textbf{GCN-SMOTE}               & 0.8617            & 0.4868            & 0.9339           & 0.7399          & 0.7763          & 0.2313          & 0.7207          & 0.2146          & 0.9184          & 0.6499          \\
\textbf{GCN-DAGAD}               & 0.8891            & 0.4792            & 0.9315           & 0.7237         & 0.7882          & 0.2741          & 0.7313          & 0.2219          & 0.9144          & 0.6507          \\
\textbf{GCN-AugAN (Ours)}          & 0.9058            & 0.4917            & 0.9492           & 0.7635          & 0.7868          & 0.2850          & \uline{0.8467}          & \uline{0.3321}          & 0.9277          & 0.7016          \\ 
\midrule
\textbf{PCGNN-DeepAll}           & 0.8716            & 0.4789            & 0.9252           & 0.7042          & 0.7681          & 0.2443          & 0.6914          & 0.1848          & 0.9335          & 0.6604          \\
\textbf{PCGNN-SMOTE}             & 0.8673            & 0.4854            & 0.9390           & 0.7403          & 0.7371          & 0.2163          & 0.7032          & 0.2021          & 0.9309          & 0.6658          \\
\textbf{PCGNN-DAGAD}             & 0.8861            & 0.4739            & 0.9367          & 0.7218          & 0.7502          & 0.2354          & 0.7163          & 0.2119          & 0.9297         & 0.6731          \\
\textbf{PCGNN-AugAN (Ours)}      & 0.8945            & 0.4815            & 0.9553           & 0.7659          & 0.7598          & 0.2487          & 0.8112          & 0.3258          & 0.9368          & 0.7176          \\
\midrule
\textbf{GDN-DeepAll}             & 0.9011            & \textbf{0.5494}   & 0.9589           & \uline{0.7821}          & 0.8760          & 0.4251          & 0.7708          & 0.3107          & 0.9718          & 0.9015          \\
\textbf{GDN-SMOTE}               & 0.9023            & 0.5426            & 0.9519           & 0.7755          & 0.7543          & 0.2335          & 0.7548          & 0.2827          & 0.9723          & \uline{0.9024}          \\
\textbf{GDN-DAGAD}              & 0.9017            & 0.5402            & 0.9528          & 0.7781          & 0.8719          & 0.4206          & 0.7815          & 0.3062          & 0.9703          & 0.8971          \\
\textbf{GDN-AugAN (Ours)}          & \uline{0.9101}            & \uline{0.5454}            & \textbf{0.9645}  & \textbf{0.7862} & \textbf{0.8927} & \uline{0.4361}          & \textbf{0.8615} & \textbf{0.3369} & \uline{0.9780}          & \textbf{0.9062} \\
\bottomrule
\end{tabular}
}
\label{Table_2_indomain_G}
\vspace{-0.15cm}
\end{table*}

\begin{table*}[h]
\centering
\caption{Model performance w.r.t AUC and AUPR on unseen testing dataset. \textbf{Bold}: best; \uline{Underline}:runner-up.}
\resizebox{0.75\linewidth}{!}{
\begin{tabular}{ccccccccccc}
\toprule
\multirow{2}{*}{\textbf{Method}} & \multicolumn{2}{c}{\textbf{Clothing}} & \multicolumn{2}{c}{\textbf{Photo}} & \multicolumn{2}{c}{\textbf{ACM}}            & \multicolumn{2}{c}{\textbf{DBLP}}           & \multicolumn{2}{c}{\textbf{CS}}             \\    \cmidrule(r){2-3} \cmidrule(r){4-5} \cmidrule(r){6-7}  \cmidrule(r){8-9}  \cmidrule(r){10-11} 
                        & \textbf{AUC}      & \textbf{AUPR}     & \textbf{AUC}    & \textbf{AUPR}    & \textbf{AUC}         & \textbf{AUPR}        & \textbf{AUC}         & \textbf{AUPR}        & \textbf{AUC}         & \textbf{AUPR}        \\
\midrule
\textbf{Dominant}                & 0.4871            & 0.0357            & 0.4989           & 0.0705          & 0.4710          & 0.0570          & 0.4937          & 0.0519          & 0.4402          & 0.0519          \\
\textbf{CoLA}                    & 0.5036            & 0.0493            & 0.5115           & 0.0677          & 0.4741          & 0.0568          & 0.4903          & 0.0512          & 0.5416          & 0.0935          \\
\textbf{SL-GAD}                  & 0.5595            & 0.0377            & 0.5410           & 0.0730          & 0.4985          & 0.0543          & 0.5019          & 0.0395          & 0.5366          & 0.0622          \\
\midrule
\textbf{OCGNN}                   & 0.6023            & 0.0455            & 0.5153           & 0.0724          & 0.5608          & 0.0765          & 0.5181          & 0.0585          & 0.5413          & 0.0711          \\
\textbf{GCN}                     & 0.6548            & 0.0792            & 0.7479           & 0.3229          & 0.6675          & 0.1542          & 0.5384          & 0.0548          & 0.6988          & 0.2536          \\
\textbf{DeepSAD}                 & 0.6686            & 0.1026            & 0.6499           & 0.1112          & 0.6414          & 0.1394          & 0.6181          & 0.0849          & 0.6724          & 0.2483          \\
\textbf{PCGNN}                   & 0.6436            & 0.0711            & 0.7615           & 0.3371          & 0.6459          & 0.1308          & 0.5329          & 0.0593          & 0.7117          & 0.2764          \\
\textbf{BWGNN}                   & 0.6399            & 0.0726            & 0.7735           & 0.3329          & 0.6831          & 0.1572          & 0.5920          & 0.0551          & 0.8035          & 0.4513          \\
\textbf{GDN}                     & 0.6629            & 0.0928            & 0.8989           & 0.4701          & 0.7592          & 0.2370          & 0.6649          & 0.1145          & 0.9166          & 0.6692          \\
\midrule
\textbf{Commander}               & 0.5895            & 0.0448            & 0.5589           & 0.0812          & 0.5923          & 0.0700          & 0.5531          & 0.0641          & 0.5561          & 0.0728          \\
\textbf{GNN-EERM}                & 0.6768            & 0.1035            & 0.6762           & 0.1299          & 0.7451          & 0.1884          & 0.7792          & 0.0926          & 0.8845          & 0.2915          \\
\textbf{GDN-MAML}                & 0.7120            & 0.1366            & \uline{0.9512}           & \uline{0.8295}          & \textbf{0.8025} & \uline{0.2573}          & 0.7813          & 0.0988          & \uline{0.9364}          & 0.6957          \\ 
\midrule
\textbf{GCN-DeepAll}             & 0.6910            & 0.0900            & 0.8107           & 0.4425          & 0.7332          & 0.2189          & 0.6465          & 0.0947          & 0.8211          & 0.3800          \\
\textbf{GCN-SMOTE}               & 0.6714            & 0.1038            & 0.8277           & 0.5196          & 0.7189          & 0.2111          & 0.6547          & 0.1023          & 0.7227          & 0.2796          \\
\textbf{GCN-DAGAD}               &0.6849            & 0.0961            & 0.8049          & 0.4357         & 0.7236         & 0.2095         & 0.6412          & 0.0915          & 0.8072          & 0.3620          \\
\textbf{GCN-AugAN (Ours)}          & 0.7157            & 0.1221            & 0.9181           & 0.6424          & 0.7714          & \textbf{0.2630} & \uline{0.8052}          & \textbf{0.2231} & 0.8305          & 0.3899          \\ 
\midrule
\textbf{PCGNN-DeepAll}           & 0.7212            & 0.0819            & 0.8142           & 0.4484          & 0.7140          & 0.2176          & 0.6216          & 0.0849          & 0.8230          & 0.3489          \\
\textbf{PCGNN-SMOTE}             & 0.6231            & 0.1261            & 0.8219           & 0.5088          & 0.6983          & 0.2061          & 0.6444          & 0.0974          & 0.7207          & 0.2645          \\
\textbf{PCGNN-DAGAD}             & 0.7061            & 0.0892            & 0.8081          & 0.4359          & 0.6955          & 0.2017          & 0.6149          & 0.0866         & 0.8037          & 0.3204         \\
\textbf{PCGNN-AugAN (Ours)}      & \uline{0.7343}            & 0.1210            & 0.9246           & 0.6601          & 0.7358          & 0.2442          & 0.7850          & \uline{0.2096}          & 0.8420          & 0.3697          \\
\midrule
\textbf{GDN-DeepAll}             & 0.7098            & \uline{0.1474}            & 0.9420           & 0.8037          & 0.7782          & 0.2422          & 0.7445          & 0.1829          & 0.9273          & \uline{0.7110}          \\
\textbf{GDN-SMOTE}               & 0.6776            & 0.1446            & 0.9372           & 0.7980          & 0.6935          & 0.1764          & 0.7081          & 0.1246          & 0.9271          & 0.6988          \\
\textbf{GDN-DAGAD}               & 0.6912            & 0.1383            & 0.9319           & 0.7971          & 0.7548          & 0.2217        & 0.7276          & 0.1504          & 0.9182          & 0.6849          \\
\textbf{GDN-AugAN (Ours)}          & \textbf{0.7941}   & \textbf{0.1603}   & \textbf{0.9855}  & \textbf{0.8765} & \uline{0.7957}          & 0.2504          & \textbf{0.8398} & 0.1677          & \textbf{0.9514} & \textbf{0.7408} \\
\bottomrule
\end{tabular}
}
\label{Table_3_new_G}
\vspace{-0.2cm}
\end{table*}

\subsection{Effectiveness Results (RQ1)}
% \vspace{-0.05cm}
$\\$
\noindent\textbf{Effectiveness on Training-Domain Graphs.}
We first evaluate the detection performance on the testing set of the training-domain data for different methods.
A fundamental requirement for GAD models is to identify as many anomalies as possible to eliminate potential losses. We present the results of AUC and AUPR in Table \ref{Table_2_indomain_G}.
We observe that (1) compared with the unsupervised detection models, the semi-supervised models (e.g., DeepSAD, GDN) basically achieve higher detection performance on the training-domain graphs; (2) leveraging the available multiple graphs for training usually brings better detection results (e.g., GDN-MAML vs. GDN);
(3) given a fixed base detection model, the variants with \textit{AugAN} usually outperform or achieve similar performance with the SOTA baseline (i.e., GDN-AugAN vs. GDN-MAML) and the counterparts equipped with other data augmentation methods (e.g., PCGNN-AugAN vs. PCGNN-DeepAll or GDN-AugAN vs. GDN-DAGAD) on the training-domain graphs.

\setlength{\abovecaptionskip}{-0.2cm}
\begin{figure*}
\begin{center}
\includegraphics[width = 0.92\linewidth]{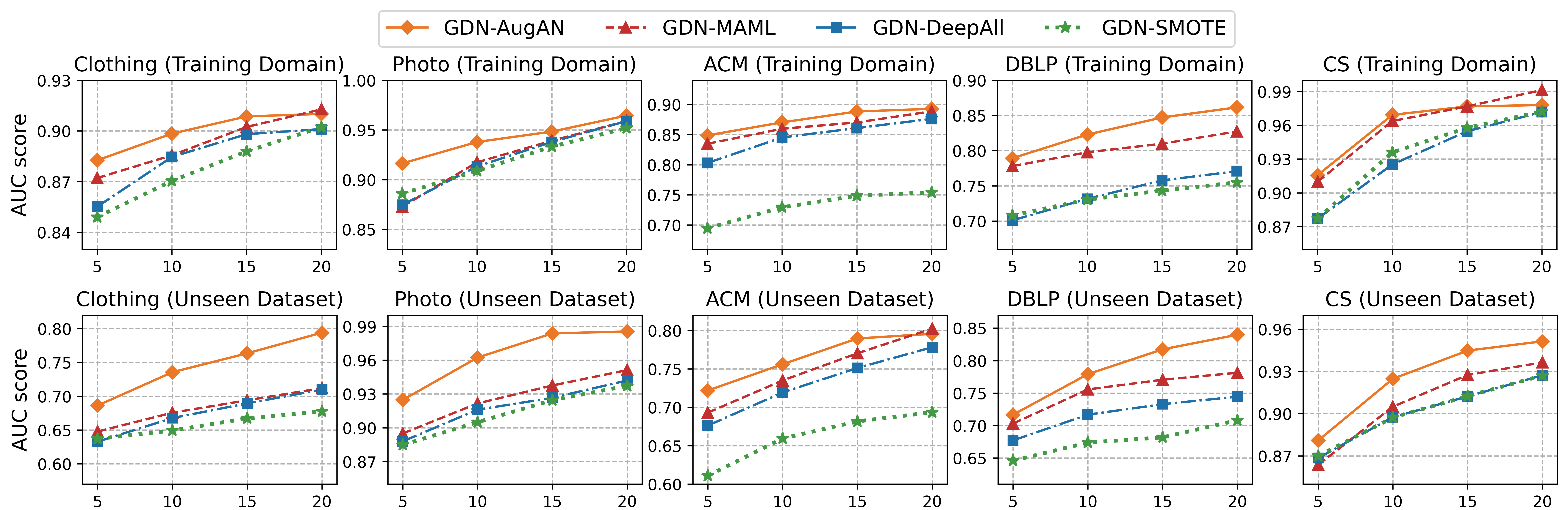}
\end{center}
\caption{AUC performance w.r.t. different number of labeled anomaly instances.}
\label{Fig3_gen_label_eff}
% \vspace{-0.4cm}
\end{figure*}

\begin{figure*}
\begin{center}
\begin{tabular}{c}
\includegraphics[width = 0.95\linewidth]{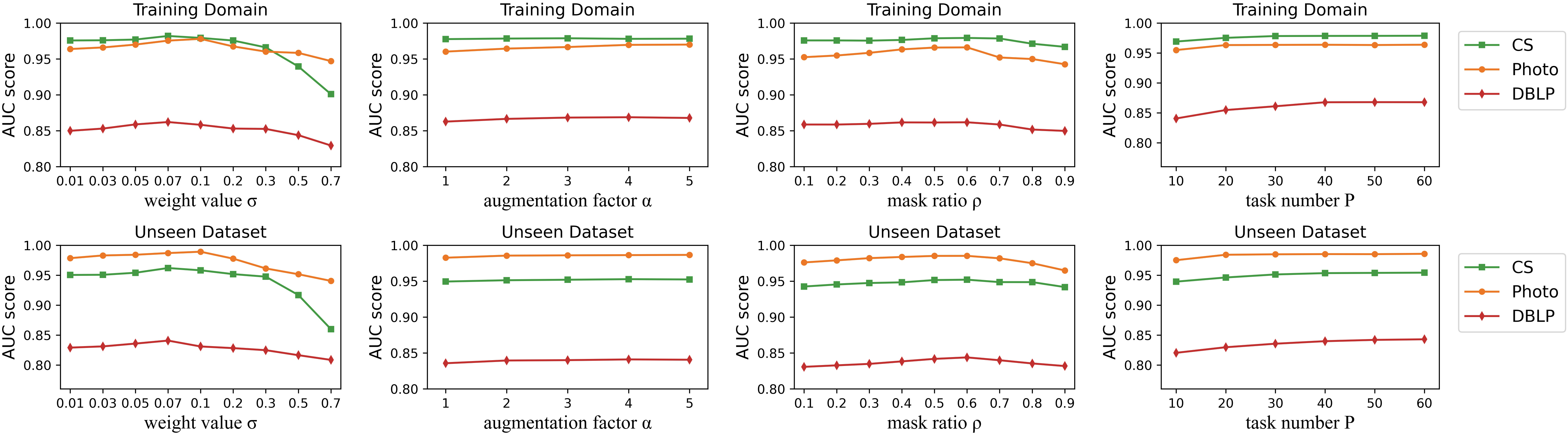}
\end{tabular}
\end{center}
\caption{AUC performance w.r.t. different hyper-parameters.}
\label{Fig4_gen_hyperparameter_analys}
\vspace{-0.15cm}
\end{figure*}

\noindent\textbf{Effectiveness on Unseen Testing Graphs.}
To examine GAD model generalizability, we also evaluate the detection performance of all the methods on an unseen testing graph (i.e., not appeared in the training).
We present the results under AUC and AUPR in Table \ref{Table_3_new_G}. 
Accordingly, we have several observations. (1) The performances of semi-supervised GAD models (e.g., GDN) generally drop severely on the unseen dataset, i.e., the semi-supervised models suffer from poor generalization.
(2) Directly adopting the trained semi-supervised GAD models on the unlabeled graph for prediction still leads to better detection results than learning an unsupervised GAD model on the unlabeled graph. It demonstrates that these semi-supervised models can still work on the unlabeled graph and boosting the generalizability of these models is important.
(3) The semi-supervised models that leverage the available multiple graphs for training basically present better generalizability than their counterparts using single-graph information for training (e.g., GDN-DeepAll vs. GDN);
(4) Given a fixed base detection model, our data augmentation method generally brings a larger performance boost than the comparative data augmentation methods. For example, GDN-AugAN outperforms GDN-DeepAll and PCGNN-AugAN outperforms PCGNN-SMOTE. 
(5) Compared with the SOTA baseline GDN-MAML, the proposed model variant with the same base detector (i.e., GDN-AugAN) achieves superior performance on four datasets while obtaining comparable performance on the ACM dataset. It may be because a relatively large number of labels are available here and the distribution shifts on the partitioned graphs of ACM are not that large, thus the anomaly augmentation does not contribute much, and the proposed episode training cannot distinguish itself from MAML. In short, the above results verify the effectiveness of our data augmentation method.

\subsection{How Labels Influence Generalization (RQ2)}
This part examines how labels influence generalization. 
In practice, it is hard to obtain a large number of labeled anomalies to train a well-generalized GAD model. However, if not providing enough labels for model learning, anomalies would escape from the detection of ``weak'' models and cause threats to benign users. Therefore, we investigate how many labels are sufficient to train a GAD model with fair generalizability. Here, we employ the GDN as the base detection model; the number of available labeled anomalies for each $\mathcal{G}^l$ varies from 5 to 20. We evaluate model performance on the test set of training-domain data and the unseen dataset. The results of some representative data augmentation methods are reported in Fig.~\ref{Fig3_gen_label_eff}.

It shows that the performances of all the GAD models increase when exploiting more labeled anomalies. Particularly, our method achieves comparable performance with baseline methods on the testing set of training-domain graphs, and consistently leads to better results on the unseen testing graph. 
We further observe that, when there exist only a few (e.g., 5) labels, adding labeled anomalies usually leads to larger performance gain than the scenarios where more labels (e.g., 15) are available.
Specifically, the results on the ACM dataset indicate that \textit{AugAN} is more likely to exceed baseline methods when only a few (e.g., 5) labels are available. It is because our proposed method \textit{AugAN} has conducted anomaly augmentation to enlarge the quantity of abnormal samples.
Considering detection effectiveness, we recommend setting a relatively large amount of labeled anomalies (e.g., 15) to achieve fair generalization.

\subsection{Stability Analysis (RQ3)}

\begin{table*}
\centering
\setlength{\abovecaptionskip}{0.02cm}
\caption{Robustness analysis results with 20 labeled anomalies. $r$ denotes the anomaly ratio. $\beta$ denotes the contamination level in training data.}
\resizebox{0.9\linewidth}{!}{
\begin{tabular}{|ccccccc|ccc|ccc|}
\hline
\multicolumn{7}{|c|}{\textbf{Data Characteristic}}                                                                                  & \multicolumn{3}{c}{\textbf{Anomaly Num (GDN-DeepAll)}} & \multicolumn{3}{|c|}{\textbf{Anomaly Num (GDN-AugAN)}}   \\ \hline
\textbf{Data}              & \textbf{nodes} & \textbf{edges} & \textbf{attributes} & \textbf{anomalies} & \textbf{$r$} & \textbf{$\beta$} & \textbf{Top 100} & \textbf{Top 200} & \textbf{Top 300} & \textbf{Top 100} & \textbf{Top 200} & \textbf{Top 300} \\ \hline
Weibo-G1 (Training Domain) & 2,195          & 23,078         & 400                 & 266                & 12.12\%    & 9.79\%     & 66               & 78               & 84               & 86               & 98               & 102              \\ \hline
Weibo-G2 (Training Domain) & 2,241          & 27,440         & 400                 & 279                & 12.45\%    & 10.27\%    & 68               & 78               & 81               & 83               & 96               & 99               \\ \hline
Weibo-G3 (Unseen Dataset)  & 2,179          & 19,370         & 400                 & 297                & 13.63\%    & -          & 86               & 148              & 204              & 99               & 178              & 237       \\      \hline
\end{tabular}
}
\label{Table_5_weibo_robust_anlysis}
\vspace{-0.3cm}
\end{table*}

\subsubsection{Hyper-parameter Analysis}
Here, we examine the influence of hyper-parameters on our proposed method. There are four major hyper-parameters in \textit{AugAN}: the weight value $\sigma$, the augmentation factor $\alpha$ in anomaly augmentation, the mask ratio $\rho$ in normal distribution augmentation, and the task number \textit{P} in the episodic training. We still employ GDN as base detection model and analyze the main hyper-parameters' influence, respectively. We evaluate model performance on the test set of training-domain data and the unseen dataset and report the results in Fig.~\ref{Fig4_gen_hyperparameter_analys}.

For the weight value $\sigma$, we observe that the detection performance firstly increases and then drops when its value is enlarged. The results may be explained by the trade-off between the quantity and the quality of labeled anomalies. In particular, raising the value of $\sigma$ helps to find more high-confidence anomalies (i.e., enlarging the label quantity) and benefit model learning. But when its value exceeds a certain threshold, more normal instances will be taken as anomalies, thus introducing noise into labels and degrading model performance.
Therefore, we recommend setting the weight $\sigma$ to a small value (e.g., 0.05-0.2) to achieve good model generalizability.
For the augmentation factor $\alpha$, the results reveal that increasing its value does not significantly boost model performance when relatively sufficient labels (i.e., 20) are available. Considering there usually lacks enough labels in real-world applications, we suggest assigning a relatively large value (e.g., 3-4) to $\alpha$ in that case. 
For the mask ratio $\rho$, we find that the performance firstly increases and then decreases as the $\rho$ value increases. This phenomenon may be explained by the trade-off between the diversity and the quantity of training data.
Specifically, when raising the value of $\rho$ (i.e., masking more normal nodes), the generated \textit{scenes} tend to present diverse backgrounds (i.e., rich diversity) and bring performance gains.
However, when the $\rho$ value is too large, the training data size severely shrinks and leads to performance drop.
Accordingly, we recommend setting the $\rho$ value to the scope of 0.4-0.7 for achieving fair generalization.
For task number \textit{P} in the episodic training, as its value increases, the model performances first increase and later become stable. We can observe that a large number of tasks helps to support the episodic training, because \textit{AugAN} simulates performing anomaly detection on the unseen dataset in the training process to find suitable model parameters for great generalization. 
Hence, setting \textit{P} a relatively large value is conducive to effectiveness. Whereas, when there exist numerous tasks, it costs longer time for model learning, especially in large graphs (refer to the running time analysis). Therefore, we recommend setting the \textit{P} value to the scope of 20-50, thus achieving a fair trade-off between effectiveness and efficiency.

\subsubsection{Robustness to Data Contamination}
Since model robustness to contaminated data has become an important concern in anomaly detection~\cite{ruff2019deep, pang2019deep}, this section investigates whether contaminated data would severely affect the effectiveness of \textit{AugAN}. 
Note that we simulate real-world scenarios where only a few labeled anomalies are available while the rest of data are unlabeled and taken as ``normal'' data for model training. Although anomalies are usually rare by nature, the existence of unlabeled anomalies in ``normal'' data may affect learning normal patterns and lead to sub-optimal performances~\cite{han2022adbench}.
Here, we aim to examine the question: can \textit{AugAN} still boost GAD model generalizability on datasets with relatively high contamination level in the training data?

Specifically, we adopt another real-world dataset (Weibo~\cite{liu2022benchmarking}) with organic anomalies and partition the graph with the aforementioned partition strategy. We keep the same experimental setting, employ GDN as the base detection model, and examine GAD model generalizability by comparing with the strong baseline method \textit{DeepAll}. We report the data statistics of each split graph and present the number of the detected anomalies in the top K (e.g., 100, 200, 300) ranking list for each graph in Table~\ref{Table_5_weibo_robust_anlysis}. 
We observe that the GAD model equipped with \textit{AugAN} can identify more anomalies than \textit{DeepAll} on both the training graphs and the unseen dataset. Specifically, the detection model with \textit{AugAN} has detected over 230 anomalies in the top 300 nodes of the ranking list in the unseen dataset, outperforming the baseline.
The results verify that the proposed data augmentation method \textit{AugAN} is robust to datasets with relatively high contamination level in training.

\begin{figure}[t]
\begin{center}
\begin{tabular}{c}
\includegraphics[width = 0.95\linewidth]{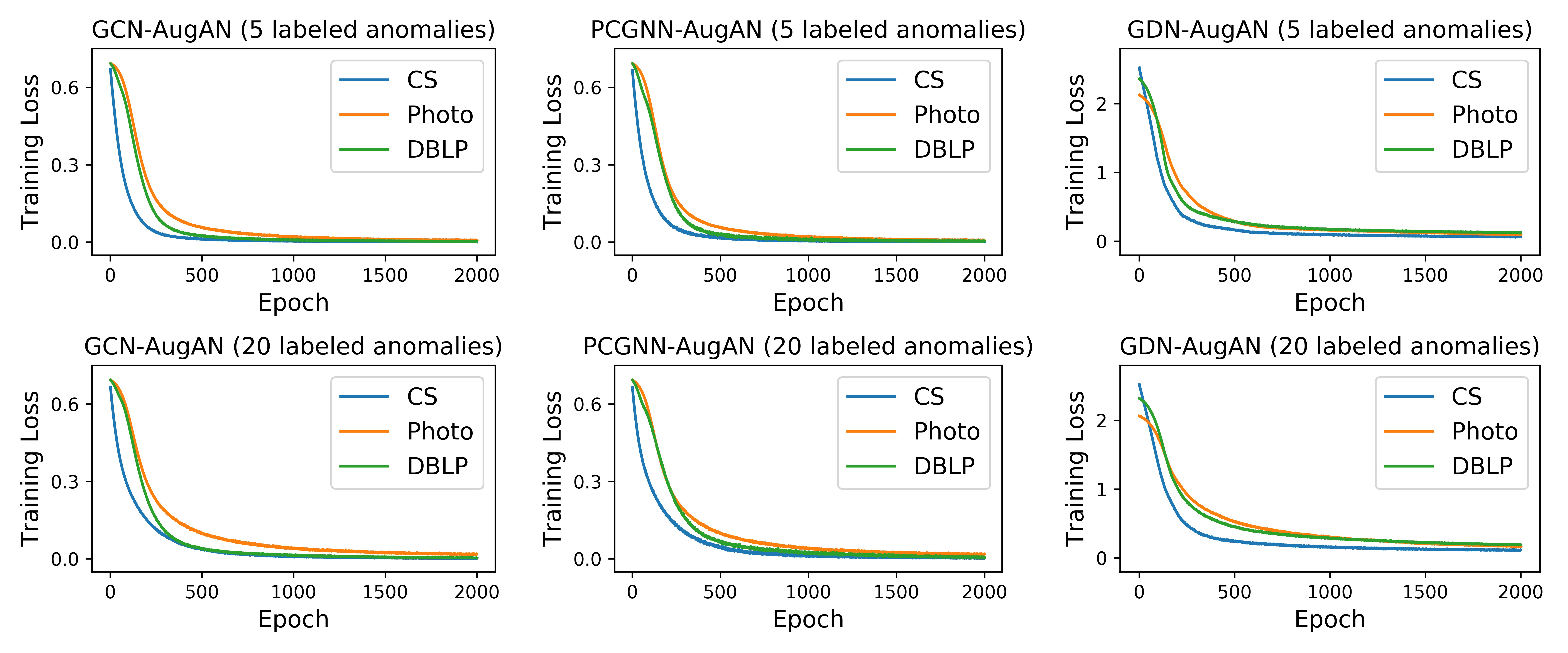}
\end{tabular}
\end{center}
\caption{Convergence analysis of different detection models equipped with the proposed data augmentation method \textit{AugAN}.}
\label{Fig_loss_curve_analys}
\end{figure}

\subsubsection{Convergence Analysis}
Since \textit{AugAN} involves customized episodic learning for boosting GAD model generalizability, it is natural to ask whether it can stably converge. In this section, we analyze the convergence to answer this question. We show the results of different detection models (GCN, PCGNN, and GDN) equipped with \textit{AugAN} on three datasets in Fig. \ref{Fig_loss_curve_analys}. The results demonstrate that the training loss converges well (i.e., the models can perform steady training) on datasets of different sizes, even if only limited labeled anomalies are provided. Therefore, in real-world applications, we think that the GAD models equipped with \textit{AugAN} can achieve steady convergence.

\subsection{Running Time Analysis (RQ4)}
\begin{figure}[t]
\begin{center}
\begin{tabular}{c}
\includegraphics[width = 0.95\linewidth]{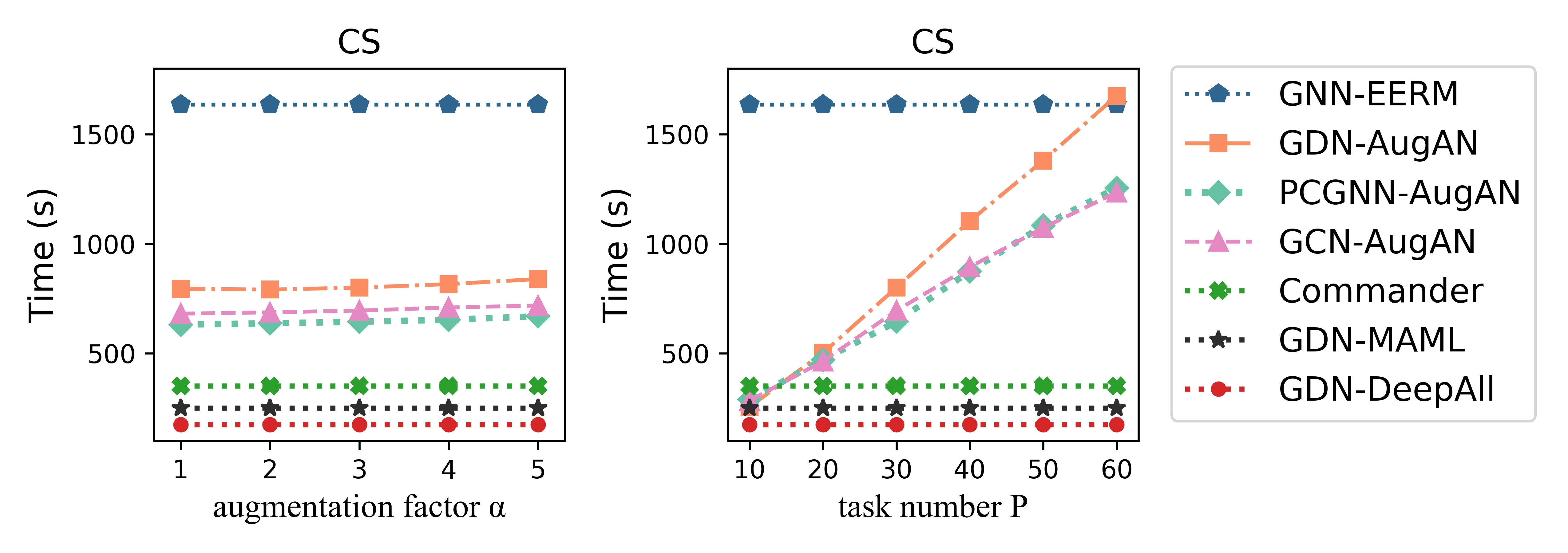}
\end{tabular}
\end{center}
\caption{Training time (seconds) of GAD models on the CS dataset.}
\label{Fig_running_time}
\vspace{-0.2cm}
\end{figure}

This section aims to examine the efficiency of our proposed data augmentation method. 
Here, we analyze the training time of the GDN-AugAN model with varying numbers of two hyper-parameters in \textit{AugAN}: augmentation factor $\alpha$, which determines the volume of the augmented anomalies, and task number \textit{P}, which affects the duration of the episodic training. 
Fig.~\ref{Fig_running_time} reports the training time performances of some representative baselines that can leverage multiple graphs for model learning. All experiments were run on an NVIDIA GTX 3090 GPU.

Based on the results, we have several interesting findings.
First, GAD models with special training mechanisms or with data augmentation usually cost more time in model learning. Specifically, Commander~\cite{ding2021cross} costs a relatively longer time than other intuitive methods (e.g., GDN-DeepAll) since Commander leverages adversarial domain adaptation that conducts model learning in a minimax game manner. Besides, GNN-EERM~\cite{wu2022handling} takes a long time on the customized datasets, because it entails multiple context generators (i.e., graph editers) that generate graph data of different virtual environments and it is adversarially trained to maximize the variance of risks from the virtual environments. 
Second, augmenting anomalies by \textit{AugAN} would not severely add burdens for model learning as the labeled anomalies are usually in limited numbers. 
Third, enlarging the task number \textit{P} in \textit{AugAN} would cost more time for GAD model training. 
Considering the fact that the trade-off between accuracy and efficiency is widely seen in the field of machine learning~\cite{yang2022region}, especially for data augmentation methods~\cite{guo2019adaptive}, we think that, by setting suitable parameters, e.g., \textit{P} equals to 30, the time efficiency of the proposed method is acceptable.

\subsection{Ablation Study and Working Mechanism (RQ5)}
\subsubsection{Ablation Study}
To better examine the contribution of each component in our proposed method \textit{AugAN}, we perform an ablation study for analysis. We respectively exclude the components, i.e., the anomaly augmentation and the normal distribution augmentation, from \textit{AugAN}. 
Note that, excluding the normal distribution augmentation means that we use the aforementioned intuitive method \textit{DeepAll}, i.e., merging all the training data, for model learning. 
Thereby, we obtain three method variants: with normal distribution augmentation only, with anomaly augmentation only, and \textit{DeepAll}. 
We still adopt the GDN as the base detection model and report its results on the unseen dataset and the testing set of training-domain data for evaluation.
Table~\ref{Table_4_ablation_label20} and Table~\ref{Table_5_ablation_label5} present the ablation study results of \textit{AugAN} with 20 and 5 labeled anomalies, respectively. 
In the tables, we present the AUC improvement over \textit{DeepAll} due to its simplicity and effectiveness for understanding the contribution of each component of the proposed \textit{AugAN} method. 

Both of the tables show that the model with either one augmentation strategy basically leads to comparable or better performance on the testing set of training-domain data. We further find an interesting phenomenon in the unseen dataset. 
When relatively sufficient labels are available, the normal distribution augmentation component contributes more to the overall performance; whereas, when there exist only a few labels, the anomaly augmentation component generally contributes more to the overall performance. 
Considering this, both components of \textit{AugAN} are necessary.

\begin{table}[]
\centering
\setlength{\abovecaptionskip}{0.02cm}
\caption{Ablation study results with 20 labeled anomalies. We report the AUC improvement over \textit{DeepAll}.}
%\textbf{Bold}: best; \uline{Underline}:runner-up.
\resizebox{0.95\linewidth}{!}{
\begin{tabular}{lcccccc}
\toprule
\multicolumn{1}{c}{\multirow{2}{*}{\textbf{Method}}} & \multicolumn{3}{c}{\textbf{Training Domain}}               & \multicolumn{3}{c}{\textbf{Unseen Dataset}}              \\
\cmidrule(r){2-4}  \cmidrule(r){5-7} 
\multicolumn{1}{c}{}                                 & \textbf{CS}     & \textbf{Photo}        & \textbf{DBLP}    & \textbf{CS}     & \textbf{Photo}   & \textbf{DBLP}    \\
\midrule
w/ Anomaly Aug.          & 0.12\%          & 0.29\%          & 1.58\%                            & 0.66\%          & 2.55\%          & 1.73\%                            \\
w/ Normal Dis. Aug.     & \uline{0.61\%}    & \textbf{0.59\%} & \uline{10.66\%}                     & \uline{2.50\%}     & \uline{4.10\%}     & \uline{11.90\%}                    \\
w/ \textit{AugAN}                   & \textbf{0.64\%} & \uline{0.58\%}    & \textbf{11.77\%}                  & \textbf{2.60\%} & \textbf{4.62\%} & \textbf{12.80\%}         \\                 
\bottomrule
\end{tabular}
}
\label{Table_4_ablation_label20}
% \vspace{-0.2cm}
\end{table}

\begin{table}[]
\centering
\setlength{\abovecaptionskip}{0.02cm}
\caption{Ablation study results with 5 labeled anomalies. We report the AUC improvement over \textit{DeepAll}.}
\resizebox{0.95\linewidth}{!}{
\begin{tabular}{lcccccc}
\toprule
\multicolumn{1}{c}{\multirow{2}{*}{\textbf{Method}}} & \multicolumn{3}{c}{\textbf{Training Domain}}               & \multicolumn{3}{c}{\textbf{Unseen Dataset}}              \\
\cmidrule(r){2-4}  \cmidrule(r){5-7} 
\multicolumn{1}{c}{}    & \textbf{CS}     & \textbf{Photo}        & \textbf{DBLP}    & \textbf{CS}     & \textbf{Photo}   & \textbf{DBLP}    \\
\midrule
w/ Anomaly Aug.      & \uline{3.44\%}    & \textbf{4.87\%} & \uline{7.86\%}     & \uline{1.02\%}    & \uline{3.52\%}    & \uline {4.42\%}    \\
w/ Normal Dis. Aug. & 2.45\%    & 4.19\%       & 6.06\%     & 0.08\%    & 0.96\%    & 1.49\%    \\
w/ \textit{AugAN}          & \textbf{4.40\%} & \uline{4.77\%}          & \textbf{12.60\%} & \textbf{1.43\%} & \textbf{4.14\%} & \textbf{5.83\%} \\
\bottomrule
\end{tabular}
}
\label{Table_5_ablation_label5}
% \vspace{-0.2cm}
\end{table}

\begin{figure*}
\begin{center}
\includegraphics[width = 1.0\linewidth]{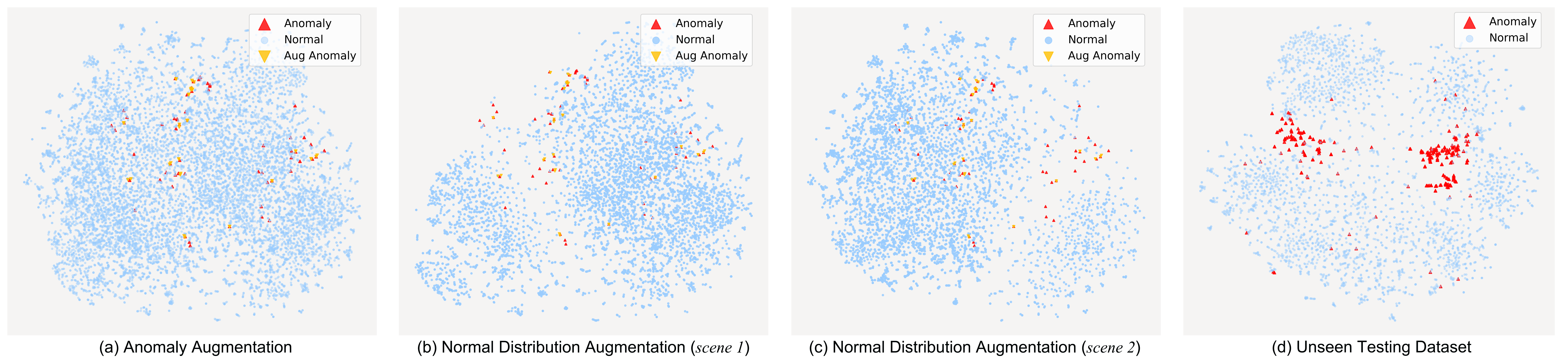}
\end{center}
\caption{Node representations visualization on the CS dataset. (a) Node representations visualization after conducting anomaly augmentation. ``Aug Anomaly" means the augmented pseudo-labels by \textit{AugAN}, which present similar characteristics to the
labeled anomalies. (b-c) Visualization results after the normal distribution augmentation. The generated multiple \textit{scenes} remain all the augmented anomalies but own diverse normal backgrounds.
(d) Node representation visualization of the unseen testing dataset. Note that, (a-c) only present the labeled anomalies, while (d) presents all the (unlabeled) anomalies.
}
\label{Fig6_TSNE_emb}
\vspace{-0.2cm}
\end{figure*}

\begin{table*}
\centering
\setlength{\abovecaptionskip}{0.02cm}
\caption{Case study on the ACM dataset. The anomalies here denote papers from rare classes (i.e., rare categories).}
\resizebox{0.95\linewidth}{!}{
\begin{tabular}{|c|c|l|l|}
\hline
\textbf{Number} & \textbf{Anomaly Type} & \multicolumn{1}{c|}{\textbf{Paper Title}}                            & \multicolumn{1}{c|}{\textbf{Abnormal Keywords}}                      \\ \hline
1               & ML                    & Bayesian learning of measurement and structural ...               & Bayesian, covariance, matrix, parametric, generalization             \\ \hline
2               & ML                    & On discriminative parameter learning of Bayesian ...  & Bayesian, classifiers, ML, Bayes, conditional, likelihood, posterior \\ \hline
3               & Robotics              & Efficient multi-robot search for a moving target                     & robotic,  indoor, environment, robot, ultra, sensor, coordination                   \\ \hline
4               & Robotics              & Adaptive compression for 3D laser data                               & laser, nonparametric, nonstationary, location, environment           \\ \hline
5               & Robotics              & Efficient base station connectivity area discovery                   & autonomy, wireless, radio, environment, signal, outdoor, sensing     \\ \hline
\end{tabular}
}
\label{Table_ACM_case_study}
% \vspace{-0.3cm}
\end{table*}

\subsubsection{Working Mechanism Analysis}
In this section, we further visualize the node representations of the training data and unseen testing dataset to reveal the working mechanism of our data augmentation method and support our claim. 
We use t-SNE~\cite{van2008visualizing} for representation visualization and depict the results on the CS dataset in Fig.~\ref{Fig6_TSNE_emb}. Note that, for abnormal samples, we only present the labeled anomalies in the training data, while presenting all the (unlabeled) anomalies on the unseen testing dataset.  
We first showcase the node representations after conducting the anomaly augmentation in Fig.~\ref{Fig6_TSNE_emb}(a). It reveals that anomaly augmentation can indeed generate pseudo-labels (i.e., augment anomalies) that present similar characteristics to the labeled anomalies. 
Then, we examine the node representations of the training data after performing the normal distribution augmentation and randomly show two of the generated \textit{scenes} in Fig.~\ref{Fig6_TSNE_emb}(b-c). We find that (1) the two \textit{scenes} indeed present a large variety on the normal background; 
(2) the distribution of the normal background and anomalies in unseen testing dataset (shown in Fig.~\ref{Fig6_TSNE_emb}(d)) is different from that of the training data. 
It may illustrate why existing semi-supervised GAD models severely degrade their performances on the unseen testing graph. Because the models may lack enough labeled anomalies for learning and also suffer from overfitting to the normal background in training. 
Accordingly, our \textit{AugAN} can both enlarge the quantity of the labeled anomalies and alleviate overfitting to the normal background by iteratively simulating the scenarios of anomaly detection on unseen data during the training process. Thereby, our method contributes to learning a well-trained GAD model that assigns anomalies (normal nodes) relatively high (low) anomaly scores.

\begin{figure}[t]
\begin{center}
\begin{tabular}{c}
\includegraphics[width = 0.8\linewidth]{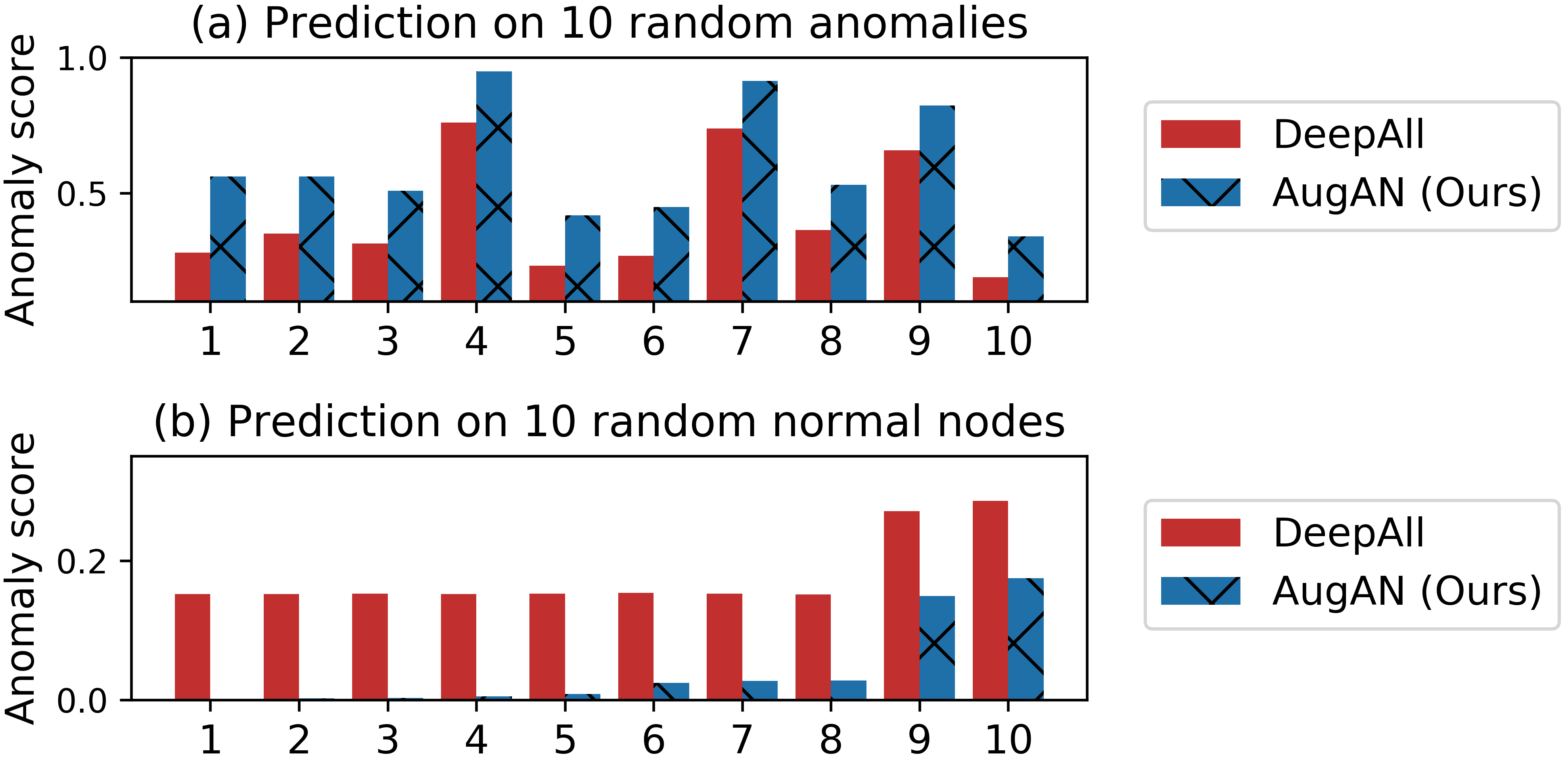}
\end{tabular}
\end{center}
\caption{Prediction on the nodes from the unseen testing dataset of CS; the adopted GAD model is respectively equipped with \textit{DeepAll} and \textit{AugAN} for training.}
\label{Fig7_score_comparison}
\vspace{-0.2cm}
\end{figure}

To verify our claim, we further compare the prediction score of anomalies and normal nodes on the unseen testing dataset.
Specifically, we examine the prediction results of a trained GAD model (e.g., GDN) that is respectively equipped with \textit{DeepAll} and \textit{AugAN} for training, and randomly select 10 anomalies and 10 normal nodes from the unseen testing dataset of CS to compare the predicted anomaly scores. We normalize the scores in (0, 1) and present the results in Fig.~\ref{Fig7_score_comparison}. 
It is clear that the GAD model trained with our method assigns anomalies higher anomaly scores, while assigning normal nodes lower anomaly scores, compared with the strong baseline method \textit{DeepAll}. 
In short, the above experiments reveal the working mechanism of our method.

\subsection{Case Study (RQ6)}
We further conduct a case study using the ACM dataset and taking GDN as the base model. As shown in Table~\ref{Table_ACM_case_study}, five sampled anomalies are listed for more specific explanations. 
The anomaly type denotes the fine-grained abnormal classes (i.e., rare categories) in the citation graph, while the abnormal keywords refer to the key information in the paper abstract. 
For example, a paper with novel keywords (e.g., radio, signal, outdoor, and sensing) is significantly different from the vast majority of papers (e.g., papers in computer vision or information retrieval) on the ACM dataset and is taken as an anomaly. The detection of such anomalies in the citation graph helps to automatically identify papers in novel research domains. 
Under this scenario, boosting the generalizability of semi-supervised GAD models can enhance the detection accuracy of these novel papers in the newly-extended area of the citation graph.

% %%%%%%%%%%%%%%%%%%%%%%%%%%%%%%%%%%%%%%%%%%%%%%%%%%%%%%%%%%%%%%%%%%%%%%%%%%%%%

% %%%%%%%%%%%%%%%%%%%%%%%%%%%%%% Related Work %%%%%%%%%%%%%%%%%%%%%%%%%%%%%%%%%%
\section{Related Work}
\subsection{Domain Generalization and Domain Adaptation}
Conventional machine learning methods generally rely on i.i.d. assumption, 
therefore, their performances usually drop sharply when there exist distribution shifts between the training data and unseen test data~\cite{wang2022generalizing}. 
To handle this issue, domain generalization (DG) and domain adaptation (DA) have attracted increasing interest recently.

DG is generally defined under the challenging setting where one or several related but different domain(s) are given, and the goal is to learn a model that can generalize to unseen test domain~\cite{wang2022generalizing}. 
To date, many techniques have been proposed for DG, including data generation~\cite{zhou2020deep, yun2019cutmix}, domain-invariant representation learning~\cite{nam2018batch, li2018deep}, and ensemble learning~\cite{thopalli2021multi}. 
Though recent works have explored DG for node classification on graph~\cite{wu2022handling}, these methods may not be suitable for GAD due to two reasons. First, in GAD, anomalies and normal data are severely imbalanced distributed and only a few labels are accessible in training. Second, methods in node classification usually implicitly rely on ``cluster” assumption~\cite{ruff2019deep}, whereas, the ``cluster” assumption may not hold on anomaly detection (i.e., anomalies may present diverse characteristics but share the same label).
Therefore, tailored methods are desired for improving generalization on GAD.

DA is a branch of transfer learning that aims to transfer the rich knowledge from related source domain(s) to boost model performance on target domain with few labels~\cite{patel2015visual}. Some methods have been proposed under the DA scenarios for anomaly detection~\cite{li2021end, ding2021cross}.
Specifically, Commander~\cite{ding2021cross} is a tailored GAD method that leverages domain adversarial learning to enhance detection accuracy on the target domain. 
A key difference between DA and the proposed generalized GAD task lies in that we aim to boost model generalizability that can generalize to (arbitrary) unseen testing data, and therefore the target domain data cannot be accessed in the training (or fine-tuning) process. In short, our task is more related to DG, compared with DA.

\subsection{Data Augmentation}
Data augmentation is a widely-used technique to handle imbalanced data issues, and many methods have been proposed for DG on other tasks, anomaly detection task, and graph structural data, respectively.
For domain generalization, data augmentation methods, which synthesize new samples for training, may alleviate the distribution shift issue, and
many related works have been conducted recently~\cite{zhou2020deep, mancini2020towards, volpi2018generalizing}. However, these data augmentation methods are primarily for images or multi-dimensional data and can hardly be applied to graph structural data.
For anomaly detection, existing data augmentation methods basically fall into two groups: augment normal data~\cite{lim2018doping, zhao2021action} or augment anomalies~\cite{rivera2020anomaly, xu2022contrastive}. 
But only a few of them can alleviate the scarcity issue of labeled anomalies and the imbalanced data issue, and be applied to attributed graph.
Specifically, DAGAD~\cite{liu2022dagad} augments training data by concatenating representations from different GNN encoders, whereas, it augments both normal data and anomalies and still cannot alleviate the imbalanced data issue.
CONAD~\cite{xu2022contrastive} leverages pre-designed rules to synthesize anomalies that fit human's prior knowledge about abnormal patterns.
However, designing the rules requires human effort, and such prior knowledge is usually hard to obtain in practice.
In addition, there also exist many node-level data augmentation methods for graphs~\cite{ding2022data, yu2022graph}, including node/edge dropping~\cite{wang2020nodeaug}, graph sampling~\cite{qiu2020gcc}, graph component synthesis ~\cite{park2021graphens}, and feature interpolation~\cite{zhao2021graphsmote, wu2021graphmixup}, but only a few methods may handle the scarcity of labeled anomalies and the imbalanced issue in GAD.
Specifically, DR-GCN~\cite{shi2020multi} adopts a GAN-based architecture~\cite{goodfellow2020generative} to generate virtual minor nodes, and GraphSMOTE~\cite{zhao2021graphsmote} performs feature interpolation to generate synthetic nodes for minority classes.
Although these data augmentation methods seem plausible to alleviate the above issues in GAD, note that, the methods are generally for node classification tasks and still rely on ``cluster” assumption~\cite{ruff2019deep}. 
As aforementioned, the ``cluster” assumption may not hold on anomaly detection, rendering them not suitable for our task. For example, directly adopting GraphSMOTE~\cite{zhao2021graphsmote} would easily sample  anomalies with different abnormal patterns, and the interpolated features may not denote abnormal patterns and are detrimental to GAD model learning.
In brief, our proposed method is different from the above methods and is customized for improving model generalizability on GAD.

\subsection{Meta Learning}
Meta-learning (a.k.a., learning to learn) is basically to learn a general model from multiple tasks and later adapt to new tasks~\cite{hospedales2020meta}. 
A common feature of meta-learning is episodic training strategy that performs model learning on the level of tasks instead of samples~\cite{wang2020generalizing}. 
Recently, many meta-learning methods have been proposed, including optimization-based (e.g., MAML~\cite{finn2017model}), metric-based~\cite{snell2017prototypical}, and model-based~\cite{santoro2016meta}. 
Specifically, GDN-MAML \cite{ding2021few} employed the MAML on GAD task and could learn anomaly patterns from auxiliary graphs to help identify anomalies on a target graph.
However, our proposed episodic training is different from GDN-MAML for two reasons. 
(1) GDN-MAML still requires a few labels on the target graph, and it highlights leveraging the labels for fast adaption, whereas, our method aims at finding suitable model parameters in the training process such that it can generalize well on (arbitrary) unseen testing data.
(2) The implementation of the episodic training exists difference. In GDN-MAML, $\mathcal{D}_{train}$ and $\mathcal{D}_{test}$ in a task are from the same graph; whereas, in our method, $\mathcal{D}_{train}$ and $\mathcal{D}_{test}$ are both randomly sampled from two different \textit{scenes} (i.e., with different background), which can simulate the scenarios of anomaly detection on unseen data in the training process.  

% %%%%%%%%%%%%%%%%%%%%%%%%%%%%%%%%%%%%%%%%%%%%%%%%%%%%%%%%%%%%%%%%%%%%%%%%%%%%%

% %%%%%%%%%%%%%%%%%%%%%%%%%%%%%% Conclusion %%%%%%%%%%%%%%%%%%%%%%%%%%%%%%%%%%
\section{Conclusion}

In this work, we find that existing semi-supervised GAD methods suffer from poor generalization issues, i.e., well-trained GAD models could not perform well on an unseen area (not accessible in training) of the graph.
Accordingly, we formally define a general research problem of generalized graph anomaly detection.
We propose a principled data augmentation method named \textit{AugAN} to boost GAD model generalizability by enlarging the quantity of labeled anomalies and enhancing the diversity of normal backgrounds and further adopting a customized episodic training strategy for learning with the augmented data.
We also release customized datasets for generalization evaluation. Extensive experiments (with 21 baselines on 5 datasets) verify the effectiveness of our method in improving GAD model generalizability.
Future works can either propose a more scalable and efficient data augmentation method for tackling the generalized GAD problem on large graphs or investigate other potentially feasible techniques, such as domain-invariant representation learning~\cite{li2018deep} and ensemble learning~\cite{thopalli2021multi}, to boost GAD model generalizability.

% %%%%%%%%%%%%%%%%%%%%%%%%%%%%%%%%%%%%%%%%%%%%%%%%%%%%%%%%%%%%%%%%%%%%%%%%%%%%%

\IEEEdisplaynontitleabstractindextext

% \fi
\IEEEpeerreviewmaketitle

\ifCLASSOPTIONcompsoc
  \section*{Acknowledgments}
\else
  \section*{Acknowledgment}
\fi

This research was supported by the grant of  DaSAIL project P0030970 funded by PolyU (UGC).

\ifCLASSOPTIONcaptionsoff
  \newpage
\fi

\end{document}